%% file: main.tex
\documentclass[10pt]{article}
\usepackage{geometry}
\usepackage{graphicx} 
\usepackage{xcolor}
\usepackage{bm}
\usepackage{mathtools}
\usepackage{amssymb,amsfonts,amsmath}
\usepackage{units}

\usepackage{tikz}
\usepackage{cite}
\usepackage{pgfplots,pgfplotstable}
\pgfplotsset{compat=newest}
\newcommand*\thankagain[1][\value{footnote}]{\footnotemark[#1]}

\usepackage{hyperref}
\hypersetup{pdfauthor={Lea Bold, Hannes Eschmann, Mario Rosenfelder, Henrik Ebel, Karl Worthmann},pdftitle={On Koopman-based surrogate models for non-holonomic robots},colorlinks=false, hidelinks}

\title{On Koopman-based surrogate models for non-holonomic robots}
\makeatletter
\let\@fnsymbol\@arabic
\makeatother
\author{Lea Bold\thanks{Optimization-based Control Group, Institute of Mathematics, Technische Universit\"at Ilmenau, Germany,
{\tt\small [lea.bold, karl.worthmann]@tu-ilmenau.de}.\newline K.\ Worthmann gratefully acknowledges funding by the German Research Foundation (DFG, project-ID 507037103).} ,
Hannes Eschmann\thanks{Institute of Engineering and Computational Mechanics~(ITM), University of Stuttgart, Germany, 
{\tt\small [hannes.eschmann, mario.rosenfelder, henrik.ebel]@itm.uni-stuttgart.de}.\newline The ITM acknowledges the support by the Deutsche Forschungsgemeinschaft (DFG, German Research Foundation) under Germany’s Excellence Strategy – EXC 2075 – 390740016, project PN4-4 “Learning from Data - Predictive Control in Adaptive Multi-Agent Scenarios” and project EB195/32-1, 433183605 “Research on Multibody Dynamics and Control for Collaborative Elastic Object Transportation by a Heterogeneous Swarm with Aerial and Land-Based Mobile Robots”.
} , 
Mario Rosenfelder\thankagain {} , 
Henrik Ebel\thankagain {} , 
Karl Worthmann\thankagain[1] 
}

\date{February 2023}

\begin{document}

\maketitle

\begin{abstract}
    Data-driven surrogate models of dynamical systems based on the extended dynamic mode decomposition are nowadays well-established and widespread in applications. Further, for non-holonomic systems exhibiting a multiplicative coupling between states and controls, the usage of bi-linear surrogate models has proven beneficial. However, an in-depth analysis of the approximation quality and its dependence on different hyperparameters based on both simulation and experimental data is still missing. We investigate a differential-drive mobile robot to close this gap and provide first guidelines on the systematic design of data-efficient surrogate models.
\end{abstract}

\section{Introduction}

Non-holonomic vehicles are of indispensible practical value in transportation and robotics. 
To automate their behavior, accurate models are key for tasks such as motion planning and model-based  
control. 
Often, in robotics, simple 
kinematic models based on first principles are employed because it can be arduous to  
take into account hardware imperfections and effects beyond kinematics, and because it fits typical cascade-type control approaches. 
An alternative are data-driven techniques, which 
need to strike a balance between data efficiency, model expressiveness, efficient and reliable numerical realizations, and, at best, should have a theoretical underpinning that may bring about beneficial theoretical properties such as quantifiable error bounds with finite data. 
With regard to these requirements, a very popular method  
is the extended Dynamic Mode Decomposition (eDMD), whose theoretical foundation is the Koopman framework. 
The Koopman operator lifts the nonlinear dynamics to linear but infinite-dimensional dynamics, which are then approximated using eDMD to generate a data-based surrogate model~\cite{BrunKutz22}. 
This approach has been recently generalized to the setting with inputs~\cite{ProcBrun18} to apply linear techniques for the controller design~\cite{BevaSosn21}. 
In this paper, we 
show, based on real-world data and hardware experiments with a non-holonomic (differential-drive) mobile robot, that and how eDMD in a Koopman framework can be used to learn a model more accurate than the nominal kinematic  
model.  
Moreover, we show how it is possible to improve data efficiency and model accuracy by incorporating physical a-priori knowledge. 

Even with  
an accurate model, controller design for non-holonomic systems remains challenging~\cite{Asto96} since, e.g., Brockett's condition is violated meaning that there does not exist a continuous time-invariant state-feedback law. 
For instance, as rigorously shown in~\cite{MullWort17, RoseEbel22}, techniques like model predictive control based on quadratic costs do not  
successfully solve the set-point stabilization problem. 
A remedy are more sophisticated schemes using structural insight, e.g., based on the homogeneous approximation and privileged coordinates, see~\cite{WortMehr15,WortMehr16,CoroGrun20,RoseEbel22}. 
This insight is key to understand whether a linear surrogate model as proposed in eDMDc suffices or a bilinear one is required~\cite{BrudFu21, FolkBurd21, OttoRowl21}. 

Extended DMD with control (eDMDc) has already been explored for robotic systems, e.g., for an inverted pendulum or a tail-actuated robotic fish~\cite{MamaCast21}, 
or within simulations for non-holonomic mobile robots~\cite{ShiKary21}. 
Even a first experimental validation of Koopman-based LQR control utilizing structural knowledge  
has been explored for a tail-actuated robotic fish~\cite{MamaCast19}. 
However, determining an optimal dimensionality of the Koopman-based surrogate model remains challenging~\cite{RenJian22}. 
A rare experimental work, in which eDMDc is applied to non-holonomic robots, can be found in~\cite{ShiKary21ACD}. 
Therein, eDMDc is used to identify a model based on simulated data using a dictionary consisting of Hermite polynomials, and the prediction of that model is also compared with the behavior of a hardware robot. 
However, the authors do not identify a model based on data from real-world hardware and, hence, only the nominal dynamics is replicated. 
Moreover, a bi-linear surrogate model seems to be advantageous as shown in~\cite{BrudFu21,FolkBurd21} on a simulated robot arm and a planar quadrotor, respectively --~a claim, which is further supported in~\cite{OttoRowl21,NuskPeit23} for control-affine systems exhibiting a state-control coupling 
since lifted linear models of finite dimension cannot capture nonlinear actuation effects inherent in many robotic systems~\cite{FolkBurd21}. 

The contribution of this manuscript is the experimental investigation of the Koopman-based, bi-linear surrogate model in simulation \textit{and} experiment, which, to the knowledge of the authors, is novel in itself and in the depth of the conducted analysis. 
In that regard, we consider the so-called one-step error to analyze and compare the prediction accuracy for various reference trajectories in dependence of the key hyperparameters like the composition of the dictionary, the amount of data points, and the control basis employed for the bilinear approach. 
In particular, we outperform nominal models using surrogate models generated from random real-world data. 

Section~\ref{sec:eDMD} recaps eDMD in the Koopman framework before the problem setup is given in Section~\ref{sec:robot}. Then, simulation and experimental results are presented in Sections~\ref{sec:simulation} and~\ref{sec:exp}, respectively, before the results are discussed and conclusions are drawn.

\bigskip\noindent\textbf{Notation}: For integers~$n,m \in \mathbb{Z}$ with~$n \leq m$, we define~$[n:m] \coloneqq \mathbb{Z} \cap [n,m]$.

\section{Recap: eDMD in the Koopman framework}\label{sec:eDMD}

We consider the 
nonlinear dynamical system governed by~$\dot{x}(t) = f(x(t))$ with a locally-Lipschitz continuous vector field~$f: \mathbb{R}^{n_x} \rightarrow \mathbb{R}^{n_x}$.
Then, for observables~$\varphi \in L^2(\mathbb{R}^{n_x},\mathbb{R})$, the Koopman operator is defined by the identify
\begin{equation}\label{eq:Koopman_operator}
    (\mathcal{K}^t \varphi)(x^0) = \varphi(x(t;x^0)) \qquad\forall\,(t,x^0) \in \mathbb{R}_{\geq 0} \times \mathbb{R}^{n_x},
\end{equation}
i.e., instead of evaluating the observable~$\varphi$ at the flow~$x(t;x^0)$ emanating from the initial condition~$x(0;x^0) = x^0$ at time~$t$, the Koopman operator propagates the observable forward in time~$\mathcal{K}^t \varphi$ and, then, evaluates the propagated observable at the initial value~$x^0 \in \mathbb{R}^{n_x}$. 
Alternatively, one may also work with the  
generator~$\mathcal{L}$ of the Koopman semigroup~$(\mathcal{K}^t)_{t \in \mathbb{R}_{\geq 0}}$, which satisfies the abstract Cauchy problem~$\dot{z}(t) = \mathcal{L}z(t)$,~$z(0) = \varphi$, see, e.g.,~\cite{SchaWort22}.
For details on DMD~\cite{Tu13} and its variants, we refer to~\cite{Schmi22} and the references therein. The connection to the Koopman framework is treated in~\cite{BrunKutz22}. Here, we restrict ourselves to a compact set~$\mathbb{X} \subset \mathbb{R}^{n_x}$, see~\cite{SchaWort22} for a detailed discussion. 

For the dictionary~$\mathbb{V} \coloneqq \operatorname{span} \{ ( \psi_j )_{j=1}^N \}$ with~$\psi_j: \mathbb{X} \to \mathbb{R}$, the data-based surrogate model of the Koopman generator using the i.i.d.\ data points~$x^{[1]}, .., x^{[d]} \in \mathbb{X}$ is given by 
\begin{align*}
    \tilde{\mathcal{L}}_{d} = \tilde{C}^{-1} \tilde{A} %
    \quad\text{ with }\quad \tilde{C} = \tfrac{1}{d} \Psi_X \Psi_X^\top\text{ and }\tilde{A} = \tfrac{1}{d} \Psi_X \Psi_Y^\top,
\end{align*}
where the matrices~$\Psi_X, \Psi_Y \in \mathbb{R}^{N \times d}$ are defined by
\begin{align*}
    \Psi_X & \coloneqq \left[ \left. \left[\begin{smallmatrix}
            \psi_1(x^{[1]}) \\ 
            : \\ 
            \psi_N(x^{[1]})
        \end{smallmatrix}\right]\right| \ldots \left| \left[\begin{smallmatrix}
            \psi_1(x^{[d]})\\
            : \\
            \psi_N(x^{[d]})
        \end{smallmatrix}\right]\right. \right], \\
        \Psi_Y &\coloneqq \left[ \left. \left[\begin{smallmatrix}
            (\mathcal{L}\psi_1)(x^{[1]})\\
            : \\
            (\mathcal{L}\psi_N)(x^{[1]})
        \end{smallmatrix}\right]\right| \ldots \left| \left[\begin{smallmatrix}
            (\mathcal{L}\psi_1)(x^{[d]})\\
            : \\
            (\mathcal{L}\psi_N)(x^{[d]})
        \end{smallmatrix}\right]\right. \right].
    \end{align*}
Note that~$(\mathcal{L}\psi_j)(x^{[i]}) = f(x^{[i]}) \cdot \nabla \psi_j(x^{[i]})$ holds for all~$(i,j) \in [1:d] \times [1:N]$. 
Since one cannot expect invariance of~$\mathbb{V}$ w.r.t.\ the approximated Koopman operator, one projects the outcome to the coordinate functions, which are tacitly assumed to be contained in the dictionary, e.g.,~$\psi_i(x) = x_i$ for all~$i \in [1:n_x]$. In the operator setting, a time shift~$\delta > 0$ is fixed and the data matrix~$\Psi_Y$ contains the entries~$\psi_j(x(\delta;x_i))$ instead of~$(\mathcal{L}\psi_j)(x_i)$.

For control-affine systems $\dot{x}(t) = f(x(t)) + \sum_{i=1}^{n_u} g_i(x(t)) u_i(t)$, there are two different options to deduce eDMD-based surrogate models. In~\cite{KordMezi18}, a linear surrogate model~$\dot{\psi} = \mathcal{L}\psi + \mathcal{B}u(t)$ (eDMDc) is proposed. To this end, the state is augmented by the control, i.e.,~$\tilde x = [x^\top\ u^\top]^\top$. 
An alternative are bi-linear surrogate models that explicitly leverage the control-affine structure, i.e., the identity~$\mathcal{L}^{u(t)} = \mathcal{L}^{0} + \sum_{i=1}^{n_u} u_i(t) (\mathcal{L}^{e_i} - \mathcal{L}^0)$, where~$\mathcal{L}^{e_i}$ is the generator for the autonomous dynamics with~$u\equiv e_i$. This yields~$\dot{\psi} = \mathcal{L}^{u(t)}\psi$, 
see, e.g.,~\cite{PeitOtto20} and the references therein. This approach seems to be preferable. 
On the one hand, it alleviates the curse of dimensionality resulting from the state augmentation in eDMDc. 
On the other hand, bilinear models seem to be superior if state-control couplings are present, i.e., one of the vector fields~$g_i$ depends on the state~$x$, see~\cite{BrudFu21,FolkBurd21,OttoRowl21,NuskPeit23}. 
For further details on the Koopman theory for control systems, see, e.g.,~\cite{BevaSosn21} and the references therein.

The approximation error can be split up into its two sources of error, i.e., the estimation~\cite{NuskPeit23} and the projection error~\cite{SchaWort22}. 
While the latter results from only finitely many observables in the dictionary~$\mathbb{V}$ and, thus, approximating the Koopman generator/operator on the respective finite-dimensional subspace, the former is a consequence of using only finitely many data points~$x^{[i]}$,~$i \in [1:d]$. 
While the convergence in the infinite-data limit also holds for eDMDc~\cite{KordMezi18convergence}, finite-data error bounds are presently only available for the bilinear approach, see~\cite{NuskPeit23,SchaWort22}.

\section{Problem Setup}\label{sec:robot}

The nominal kinematics of the 
differential-drive robot is given in terms of the driftless control-affine system
\begin{align}\label{eq:nominal_kinematics}
    \dot{x} (t) = \begin{bmatrix} \cos \theta(t) \\ \sin \theta(t) \\ 0 \end{bmatrix} v(t) + \begin{bmatrix} 0 \\ 0 \\ 1 \end{bmatrix} \omega (t), 
\end{align}
$x(0)=x^0$. The state~$x = [ x_1\ x_2\ \theta ]^\top \in \mathbb{X} \subset \mathbb{R}^3$ consists of its position~$[ x_1\ x_2]^\top$ in the plane and its orientation~$\theta$ measured relative to the~$x_1$-axis. 
Nominally, it is assumed that the robot can instantaneously attain any admissible translational velocity~$v$ in forward direction and angular yaw velocity~$\omega$, so that these act as the system's control input~$u= [ v\ \omega ]^\top\in\mathbb{U}\subset\mathbb{R}^2$, where~$\mathbb{U}$ is compact, convex, and~$0\in\text{int}(\mathbb{U})$.
In general, the nominal kinematics does not perfectly describe the dynamics of the physical robot since inertia effects, motor dynamics, and manufacturing imperfections are not accounted for. 
From a mechanical point of view, the dynamics~\eqref{eq:nominal_kinematics} 
describe the kinematics of a differential-drive mobile robot in the plane under the common assumption that the wheels roll without slipping with the wheel-floor contact point sticking perfectly to the ground, preventing instantaneous lateral motions of the robot and thereby giving rise to a non-holonomic kinematic constraint. 
A physical robot with such a kinematic setup is employed throughout this contribution.  
On the nominal kinematic level, the robot's configuration is completely described by means of its pose, hence it is sufficient to formulate the observables based on~$x$.
Thus, in general, the learning procedure from Section~\ref{sec:eDMD} receives as data recorded pairs of states and corresponding successor states, but not any prior information on the dynamics of the robot.
However, in Sec.~\ref{sec:simulation}, we show how some mechanical prior knowledge can be incorporated, e.g., when choosing the observables of the dictionary~$\mathbb{V}$.

\section{Simulation results}\label{sec:simulation}
\begin{figure}
    \centering%
    \input{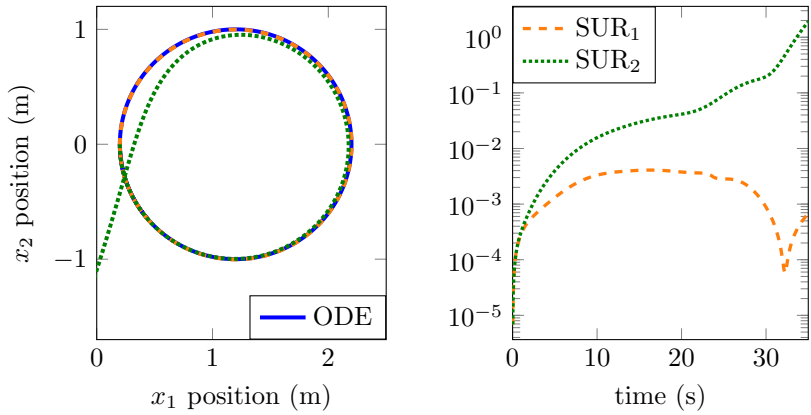}
    \caption{Results from two Koopman-based surrogate models based on first-principles data, with the trajectory emanating from~$x^0 = [0.2\ 0\ -\pi/2 ]^\top$ on the left, and the norm of the prediction error on the right.}
    \label{fig: modelbased circle}%
\end{figure}

In this section, eDMD is applied to the simulated, nominal non-holonomic robot. 
First, we generate i.i.d.\ random data matrices~$X_i \in \mathbb{R}^{n_x \times d}$,~$i \in \lbrace 0,\dots,n_u\rbrace$, with~$n_x = 3$ and~$d=10000$ data points each, where each column is in the set~$\mathbb{X}$ and serves as an initial condition for the dynamical system~\eqref{eq:nominal_kinematics}. 
Each data point contained in~$X_i$ is simulated forward 
$\delta = 0.02\,\textnormal{s}$ with the Runge-Kutta method of fourth order  
using a specific constant control input~$u_i$. 
For~$i=0$, the latter is chosen to~$u_0 = 0$. 
For~$i>0$, it is selected to be the~$i$th vector of a basis~$B$ of~$\mathbb{R}^{n_u}$. 
Here, with~$n_u=2$ and the basis~$B=\lbrace u_1, u_2\rbrace$, this yields the matrices~$Y_0$,~$Y_1$, and~$Y_2$ containing in each column the successor states of the states in~$X_0$,~$X_1$, and $X_2$ for the inputs~$u_0$,~$u_1$, and~$u_2$, respectively. 
Nominally, the system is free of drift, i.e.,~$Y_0 = X_0$. 
In simulations, different from experiments, it is possible to choose~$X_0 = X_1 = X_2$,  
which is done in this section. 
In the following, the dictionary~$\mathbb{V}$ is spanned by the monomials of~$x_1$,~$x_2$, and~$\theta$ of degree less or equal than 
7, which yields 
$N=120$ 
observables in total, yielding the set of observables~$\mathbb{O}_{120}$. 
By lifting the matrices~$X_i, Y_i$,~$i \in [0:n_u]$, with those observables, the matrices~$\Psi_{X_i}, \Psi_{Y_i}$ are computed, see Section~\ref{sec:eDMD}.  
Now, an approximation of the Koopman operator for step size~$\delta$  
is computed by 
$K_i^{\delta} = ( 
     (\Psi_{X_i} \Psi_{X_i}^\top )^{-1} 
     \Psi_{X_i} \Psi_{Y_i}^\top )^\top$,~$i \in [0:n_u]$. 
Using the bilinear  
approach, we approximate the Koopman operator  
for a control value~$u \in \mathbb{U}\subset\mathbb{R}^{n_u}$ by 
$K_u^{\delta} = K_0^{\delta} + \sum_{i = 1}^{n_u} g_i \cdot \left(K_i^{\delta} - K_0^{\delta}\right)$
for factors~$g_i$,~$i \in[ 1 : n_u ]$, which, here, solve the linear system~$g_1 u_1 + g_2 u_2 = u$.
There are two different ways to use the approximated Koopman operator to obtain the approximate values of the coordinates at a time step~$k>0$. 
In the first surrogate model variant proposed in~\cite{PeitOtto20}, subsequently referred to as SUR$_1$, one projects after 
each  
time step, i.e.,~$x_j[k] = ( \mathcal{K}^{\delta}_{u[k-1]} \Psi ( x[k-1] ) )_{j}$ for~$j \in [1:n_x]$, 
with the number inside square brackets denoting the time step, where one step is of duration~$\delta$. 
Between time steps, the new values of the observables are calculated based on the new coordinate values. 
In the second variant, called SUR$_2$ in the following, one projects once at the end, i.e.,
$x_j[k] =  (( \prod_{i=0}^{k-1} \mathcal{K}^{\delta}_{u[i]} ) \Psi(x^0) )_{\! j}$. 
To analyze their influence, Fig.~\ref{fig: modelbased circle} shows prediction results for the two variants and, as a reference, the result of the time integration of the nominal kinematic model using the Runge-Kutta method of fourth order. 
\def\lineWidthSUR{1.5}%
\def\lineWidthSURError{1.25}%
\definecolor{ODE}{RGB}{0,0,255}%
\definecolor{SUR1}{RGB}{255,127,14}%
\definecolor{SUR2}{RGB}{0,128,0}%
In the depicted scenario, the control input is set to the constant value~$u \equiv \begin{bmatrix} 0.2 & 0.2 \end{bmatrix}^\top$, i.e., the robot will move in a circle and the basis is~$B = \{ e_1, e_2\}$ for the unit vectors~$e_1, e_2 \in \mathbb{R}^{n_u}$. 
As can be seen, SUR$_1$ leads to a trajectory whose error remains comparatively small over the whole trajectory. For the model SUR$_2$, however, we receive a trajectory that visibly deviates from the reference after about one quater of simulated time, which can also be seen in the error plot. 
In the second half of the simulation, the prediction based on SUR$_2$ becomes increasingly inaccurate and quickly unusable.  
Consequently, from now on, we will only use SUR$_1$ for Koopman-based surrogate models.

\begin{figure}
    \centering
    \input{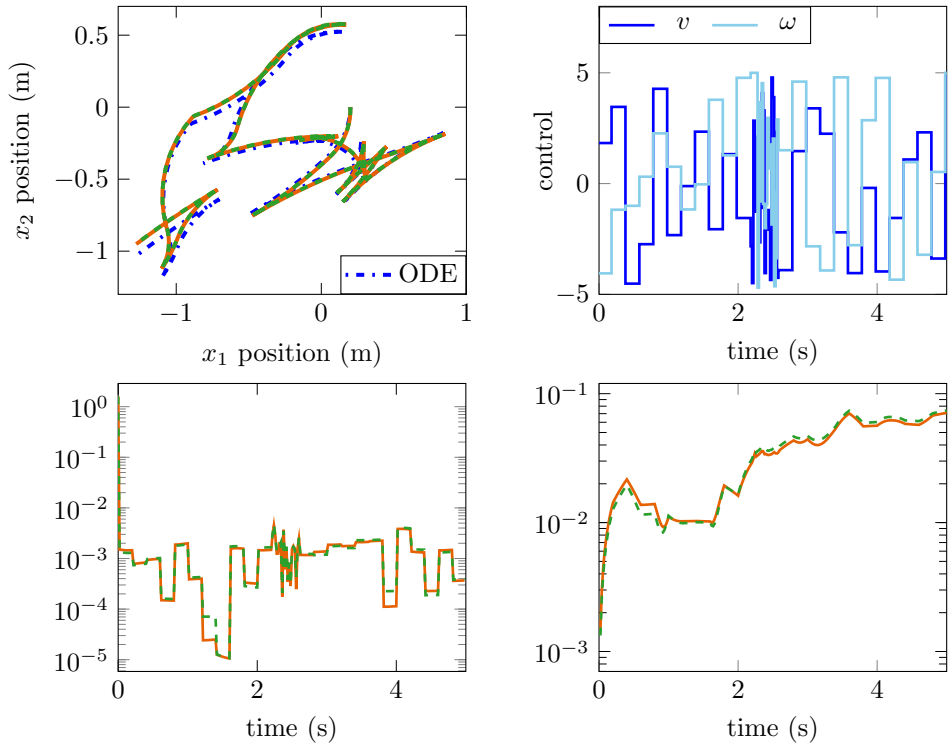} 
    \caption{Results using SUR$_1$ and the basis~$B_1$~\eqref{plot:Sur1B1} or~$B_2$~\eqref{plot:Sur1B2}. From left to right, top to bottom, the resulting trajectories in the motion plane, the applied control values, the one-step prediction errors, and total error norms are shown.}
    \label{fig: firstprinciple}
\end{figure}

The basis employed for~$\mathbb{R}^{n_u}$ need not consist of unit vectors. 
In the following, we use the bases~$B_1 = \{ [0.2\ 0]^\top, [0\ 2]^\top \}$ and~$B_2 = \{ [0.2\ -0.4]^\top, [0.2\ 0.6]^\top \}$ instead. 
Basis~$B_1$ contains scaled variants of the unit vectors that, in absolute value, fit better to the usual operating points of the employed hardware robot; for instance, it cannot attain translational velocities of~$1\,\textnormal{m}/\textnormal{s}$. 
Still, training with~$B_1$ only captures the robot driving a straight line or rotating on the spot. 
In contrast, to study the influence of the usage of different training motions for learning, the inputs contained in~$B_2$ let the robot drive arcs of different radii.  
In Fig.~\ref{fig: firstprinciple}, the results for those two bases are illustrated. 
\definecolor{ODEcolor}{RGB}{0,0,255}
\definecolor{SUR1B1color}{RGB}{230,97,1}  
\definecolor{SUR1B2color}{RGB}{44,160,44} 
\definecolor{lightblue}{RGB}{135,206,235}
\def\lineWidthStep{1.0}
\def\lineWidthPlane{1.5}
\def\heightODEtop{5.4cm}
\def\heightODEbottom{5.4cm}
\def\yBottomPlots{-5.0cm}
In the plotted scenario, the same random control sequence~$u$ is applied to the models. 
Once again, the prediction results of the surrogate models are compared to time integrations of the nominal model, which is used as a reference.  
In addition to the error norm, the one-step prediction error is considered. 
To calculate the latter, in each time step, starting from the same reference value, the following time step is predicted using the model of choice and the result is compared with the corresponding, subsequent value of the reference. 
As the results in Fig.~\ref{fig: firstprinciple} show, using random control values leads to a higher error than using the constant control input from Fig.~\ref{fig: modelbased circle}, motivating the subsequent analysis using test trajectories where a wider variety of inputs are applied. 
Moreover, here, the difference between the two bases is negligible. 
However, it is not a priori clear whether the latter also holds when using data from an imperfect hardware robot. 
Hence, real-world data is considered subsequently. 

\section{Experimental results}\label{sec:exp}
We use a custom-built mobile robot as depicted on the right of 
Fig.~\ref{fig:trainingData}. 
Its pose is tracked by an external tracking system consisting of five Optitrack Prime 13W cameras. 
The robot receives its inputs, the desired forward translational velocity and the desired angular yaw velocity, wirelessly. 
On-board software kinematically calculates the angular velocities of the wheels corresponding to the inputs under the assumption of rolling without slipping. 
Two independent PID controllers operating at a frequency of~$100\,\textnormal{Hz}$ control the motors so that the wheels quickly attain the desired angular velocities.  
Naturally, 
due to imperfections, the actual robot velocities may not match the sent ones. 
The time step is set to~$\delta = 0.1\,\textnormal{s}$ subsequently. 

\subsection{Data Generation}
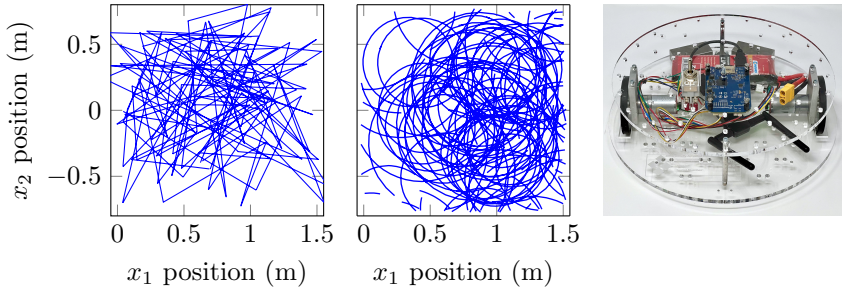
\begin{figure}
    \centering
    \input{trainingData.tex}
    \caption{From left to right, training trajectories used to generate the samples for the input bases~$B_1$ and~$B_2$, and a photograph of the employed type of custom-built robot are shown.}
    \label{fig:trainingData}
\end{figure}

Generating uniformly distributed training samples is possible by driving the robot to each corresponding point in the state space individually, applying one of the~$n_u$ inputs, and potentially driving back to that point to apply another input. 
However, this way of generating training data is notoriously time-inefficient. 
The more efficient procedure used in this paper works as follows. 
For the considered robot, holding any input for several time steps nominally results in a circular motion with the radius being determined by the quotient of the translational and angular velocities.  
The basis vectors of~$B_1$, which consist of driving in a straight line and turning on the spot, correspond to circles with infinite and vanishing radii, respectively. 
Therefore, slightly different sampling strategies for the two input bases~$B_1$ and~$B_2$ are used.
Starting from an initial position on the admissible motion plane~$\mathbb{P}=[0.0, 1.5]\,\textnormal{m}\times[-0.75, 0.75]\,\textnormal{m}$  with~$\mathbb{X}= \mathbb{P}\times \mathbb{R}$, a new point is drawn i.i.d.  
For~$B_1$, the robot turns using the corresponding input of the input basis until it faces this generated point. 
In order to collect as many data points as possible, the robot does at least one full rotation. 
Subsequently, the robot drives in a straight line toward this generated point. 
This way, the necessary input of~$B_1$ is held for several time steps, generating training samples along the way, making the procedure very time efficient. 
This procedure is repeated until a sufficient amount of training data is generated.
Due to the reasons stated above, for the input basis~$B_2$, the sampling strategy is adjusted slightly.
The robot, again, turns and drives towards the uniformly randomly generated point. 
Then, each time alternating between the two basis vectors of~$B_2$, the inputs are applied either until a full circle is driven or until the nominal state prediction of the robot leaves~$\mathbb{X}$. 
Generally, while time efficient and effective, this way of generating samples does not lead to a perfectly uniform distribution.
In Fig.~\ref{fig:trainingData}, some of the trajectories used during the data generation 
are depicted.\footnote{We do not immediately apply the full magnitude of the basis vectors for a more controlled behavior. 
Data points during acceleration and deceleration are not used since they do not correspond to any input basis vector. 
In case the random point is too close to the previous one to reach the desired speed, the point is discarded and a new one is generated so that enough data can be gathered in-between.}
Another practical consideration concerns the measurement of the robot's orientation. 
The optical tracking system steadily continues the angular measurements such that the orientation angle may lie outside of~$(-\pi, \pi]$. 
While it would be possible to use the raw data for training, instead, we leverage the periodicity of the orientation.%
\footnote{Each orientation in~$X_i$,~$i\in\{0,1,2\}$, is shifted to its equivalent value within~$(-\pi, \pi]$. 
The entries of~$Y_i$ are shifted by the same amount as the corresponding entries of~$X_i$. 
However, after that, some orientations in~$Y_i$ may still lie outside of~$(-\pi, \pi]$, namely if the angle left the interval between the sampling instants. 
The matrices with shifted entries are then used to compute the surrogate model. 
Before each evaluation of the model, 
the orientation is shifted to~$(-\pi, \pi]$. 
Subsequently, the output is then shifted back, resulting in the surrogate model being periodic (but not necessarily continuous) in the orientation.}

\subsection{Results}
Two main scenarios are considered.   
In the first scenario, the robot follows an~$\infty$-shaped trajectory. 
As the terminal and initial velocities are zero, at the start as well as the end of the trajectory, the speed is increased or decreased linearly to obtain a smoother motion. 
In the second scenario, the robot shall follow a square-shaped trajectory. To that end, the robot drives trapezoidal velocity profiles on each edge of the square with a top speed of about \unit[0.2]{m/s}. At each corner, the robot makes a quarter counter-clockwise turn with a maximum absolute angular velocity of \unit[1.0]{rad/s}, with the angular velocities being increased or decreased linearly. 

\begin{figure}
    \centering
    \input{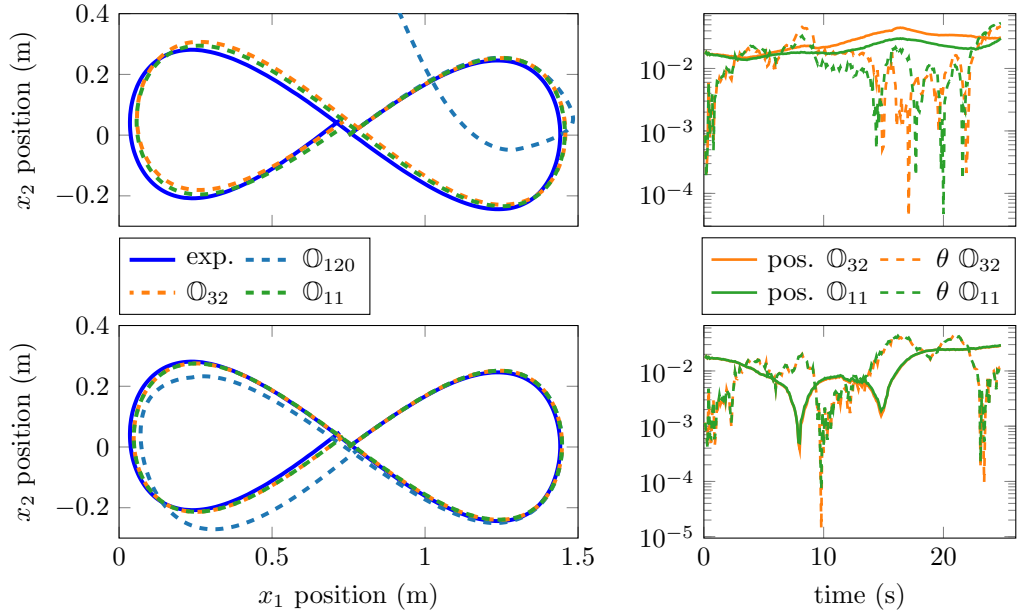} 
    \caption{Results using~$B_1$ (top) and~$B_2$ (bottom) based on real data, where trajectories for different sets of observables are compared with the result of an experiment run. Absolute errors are depicted on the right, independently for position (pos., norm) and orientation ($\theta$). }
    \label{fig: all observables}
\end{figure}

First, we look at Koopman-based models in which we do not incorporate further a-priori knowledge. 
The training data was generated as described above for the constant controls contained in the bases~$B_1$ or~$B_2$, for which~$4626$ or~$5182$ training data points were recorded, respectively.  
Because of the results from Section~\ref{sec:simulation}, only the Koopman-based surrogate model with projection in each step (SUR$_1$) is employed.   
Results for the~$\infty$-shaped trajectory can be seen in Fig.~\ref{fig: all observables}, where for the two bases~$B_1$ and~$B_2$ as well as for different observable sets, the resulting predicted trajectories are plotted on the left-hand side and the absolute errors are compared on the right-hand side. 
The errors are measured relative to a representative lap of the hardware robot. 
Due to imperfections, when supplied with inputs that should lead to a perfect~$\infty$-trajectory for the nominal kinematics, the real robot's trajectory is not of perfect shape. 
Three different surrogate models differing in their dictionaries are considered. 
Firstly, the set of observables~$\mathbb{O}_{120}$ from~Section~\ref{sec:simulation} is used. 
Secondly, in~$\mathbb{O}_{32}$, compared to~$\mathbb{O}_{120}$, we exclude monomials for which~$x_1$ and~$x_2$ have a degree larger than~$1$, yielding~$32$ observables in total. 
Finally, we further reduce the number of observables by omitting monomials where~$x_1$ or~$x_2$ appear multiplied with~$\theta$, leading to~$\mathbb{O}_{11}$ with~$11$ observables. 
This is motivated by the physical insight that the robot's dynamics is translation invariant,  
so it is interesting to see whether incorporating this knowledge improves model quality. 
In that regard, as can be seen in the upper part of Fig.~\ref{fig: all observables}, the predictions of the surrogate model using~$B_1$ with~$\mathbb{O}_{120}$ follow the reference rather well for some time but then completely deviate and even leave the experiment area. 
The paths for~$\mathbb{O}_{32}$ and~$\mathbb{O}_{11}$, however, are nearly indistinguishable and follow the reference well; only the error plot suggests that~$\mathbb{O}_{11}$ might perform a bit better. 

\begin{figure}
    \centering%
    \input{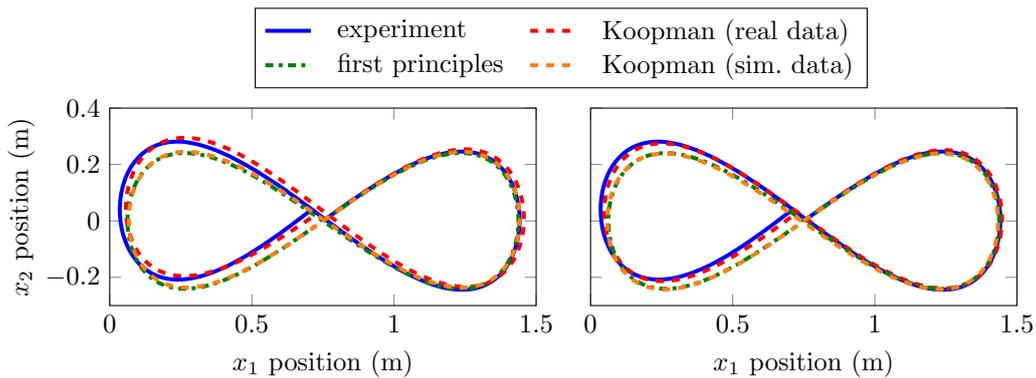}%
    \caption{Comparison of the surrogate models using~$B_1$ and~$B_2$ with models based on first principles and on data generated from a first principles model.}
    \label{fig: mixed}%
\end{figure}

To study the influence of the input basis, the same scenario is plotted in the bottom part of Fig.~\ref{fig: all observables} for basis~$B_2$. 
\definecolor{dic120}{RGB}{31,119,180}%
\definecolor{dic32}{RGB}{255,127,14}%
\definecolor{dic11}{RGB}{44,160,44}%
\definecolor{reference}{RGB}{0,0,255}%
\def\linewidthEight{1.5}%
\def\linewidthError{1.0}%
Again, the trajectories for~$\mathbb{O}_{32}$ and~$\mathbb{O}_{11}$ are very close to each other, even in the error plot. 
However, with~$B_2$, using the observables~$\mathbb{O}_{120}$ results in a trajectory that is close to the reference for much longer before the error becomes visible. 
Therefore, these results seem to suggest that using the basis~$B_2$ is a lot better if the set of observables~$\mathbb{O}_{120}$, which does not use any physical insight, is used and slightly better if~$\mathbb{O}_{32}$ and~$\mathbb{O}_{11}$ are employed, which partly or fully presume translational invariance. 
Due to these findings, we subsequently use the set of observables~$\mathbb{O}_{11}$ since it seems to yield the best predictions but, due to less elements, is also the most computationally efficient. %
In particular, as Fig.~\ref{fig: mixed} shows, the surrogate models using~$\mathbb{O}_{11}$ beat the predictions of the nominal model as well as (naturally) of the surrogate model from Section~\ref{sec:simulation}. 
\definecolor{KoopData}{RGB}{255,0,0}%
\definecolor{RungeKutta}{RGB}{0,128,0}%
\definecolor{KoopFP}{RGB}{255,127,14}%
\definecolor{reference}{RGB}{0,0,255}%
\def\linewidthEightB{\linewidthEight}%
\def\linewidthError{1.0}%
\def\heightEight{4.2cm}%

\begin{figure}
    \centering%
    \input{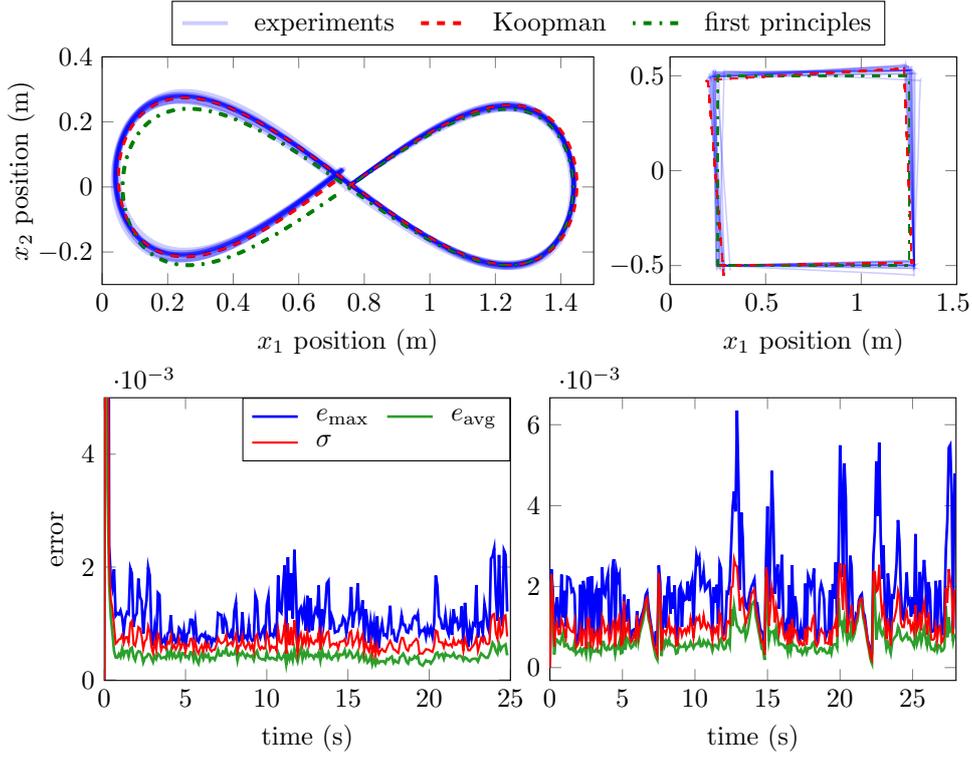}%
    \input{figures/tikz/mean_max_8}%
    \caption{Comparison of the surrogate model using~$B_2$ with 15 experiment runs, where~$e_{\textnormal{max}}$,~$e_{\textnormal{avg}}$, and~$\sigma$ denote maximum, average, and standard deviation of the error norms, respectively.}  
    \label{fig: data B2 ref8}%
\end{figure}

From now on, due to the beneficial performance, if not stated otherwise, we employ the basis~$B_2$ with the observables~$\mathbb{O}_{11}$ and compare the prediction performance of the corresponding surrogate model with the nominal model and experiment runs.  
In particular, we include~$15$ experiment runs for each scenario since subsequent experiment realizations generally do not yield identical results due to disturbances, meaning that perfect prediction performance is impossible. 
Results for the~$\infty$-trajectory and for the square-shaped trajectory are plotted in Fig.~\ref{fig: data B2 ref8}. 
From the trajectory plots in the upper part, it becomes evident that the Koopman-based prediction outperforms the nominal model, better representing the systematic skewedness of the physical robot's trajectories. 
Similarly, in the lower part, the minimum, maximum and average Euclidean norms of the error  
between Koopman-based prediction and the family of hardware robot trajectories show that prediction quality is consistently good. 
    
\definecolor{green}{RGB}{0,128,0}
\definecolor{Koopman}{RGB}{255,0,0}
\definecolor{RungeKutta}{RGB}{0,128,0}
\definecolor{reference}{RGB}{0,0,255}
\def\linewidthEightC{\linewidthEight}
\def\linewidthErrorC{1.0}
\def\linewidthErrorStdVar{0.8}
\definecolor{gray}{RGB}{128,128,128}
\definecolor{orange}{RGB}{255,165,0}
\definecolor{MaxError}{RGB}{0,0,0255}
\definecolor{AvgErr}{RGB}{44,160,44}%
\def\opacityRef{0.2}
\def\heightRealData{4.6cm}
\def\heightRealDataError{5.33cm}

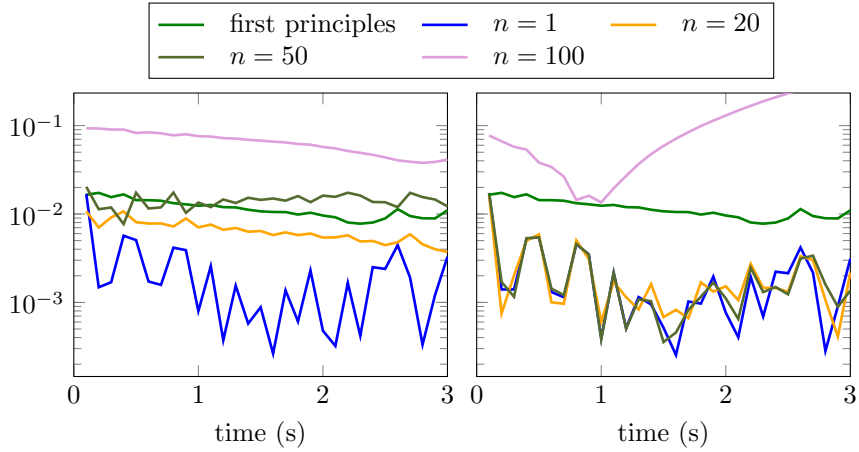
\begin{figure}[t!]
    \centering%
    \input{figures/tikz/lessdata_b1_new}%
    \caption{Average one-step prediction errors for surrogate models using~$B_1$ (left) and~$B_2$ (right) when only using every~$n$th data point.}%
    \label{fig: lessdata}%
\end{figure}

A remaining concern is data efficiency. 
Hence, subsequently, training data points are removed systematically to obtain smaller training data sets. 
The original training data was generated by choosing, for~$B_1$,~$m_1 = 50$ and, for~$B_2$,~$m_1 = 39$ random initial conditions in~$\mathbb{X}$. 
The relevant trajectory pieces driven for each sampled point are all of different lengths, e.g., depending on the distance to the boundary of~$\mathbb{X}$. 
To systematically reduce the number of data points, first, these lengths are unified by taking the length of the shortest trajectory, which, here, consists of~$m_2 = 20$ steps, and discarding the data points beyond that for each trajectory segment. 
This leads to a new training data set of~$m_1 \cdot m_2 = 1000$ or~$780$ per basis. 
Then, every~$n$th data point,~$n \in \{1, 20, 50, 100\}$, is used to create training data sets of lower cardinality.  
Using~$\mathbb{O}_{11}$,  the different resulting surrogate models' average one-step prediction errors w.r.t.\ the~$15$~recorded trajectories in the~$\infty$-scenario are given in Fig.~\ref{fig: lessdata}. 
\definecolor{RungeKuttaError}{RGB}{0,128,0}%
\definecolor{n1}{RGB}{0,0,255}%
\definecolor{n20}{RGB}{255,165,0}%
\definecolor{n50}{RGB}{85,107,47}%
\definecolor{n100}{RGB}{221,160,221}%
\def\lineWidthErrorN{1.0}%
\def\heightErrorN{5.34cm}%
These show that basis~$B_2$ seems to be more data efficient since the errors remain lower in data-sparse settings.  
Moreover, comparatively small training data sets can suffice in this scenario to achieve one-step prediction errors that are consistently smaller than that of the nominal model, especially when using basis~$B_2$. 

\section{Summary and Outlook}
This contribution showed with a detailed analysis that a bilinear eDMD approach in the Koopman framework can be a very powerful data-driven modeling tool in mobile robotics.  
Even with a modest amount of data and a calculation time in the second range, 
the approach can be used to learn a dynamical model that is on average more accurate in predictions than the common nominal kinematic model of a differential-drive robot. 
Moreover, we have seen that and how physical a-priori knowledge can be successfully incorporated into the model, which is interesting beyond the considered application scenario. 
In particular, we have shown how the dictionary of observables can be modified to account for translation invariance. 
Still, there are many remaining topics that we will cover in subsequent research. 
This includes data-driven modeling that strives to include second-order effects such as actuator dynamics and inertia, complicating especially practical considerations such as measuring and sampling of training data. 
Similarly, we will look at non-holonomic vehicles of higher degree of non-holonomy. 
Moreover, we intend to use the learned models for data-based predictive control.
\\

\noindent\textbf{Acknowledgement}: We sincerely thank Manuel Schaller (TU Ilmenau) for his support w.r.t.\ implementation details and fruitful discussions, which improved our manuscript.

\bibliographystyle{IEEEtran}
\bibliography{references_Koopman_robotics, references_nonholonomic_robot}

\end{document}

%% file: trainingData.tex
\begin{tikzpicture}

\newlength{\dataWidth}
\setlength{\dataWidth}{28mm}
\begin{axis}[%
height=\dataWidth,
width =\dataWidth,
at={(0cm,0cm)},
scale only axis,
tick label style={/pgf/number format/fixed},
xtick distance=0.5,
xmin=-0.05,
xmax=1.55,
ymin=-0.8,
ymax=0.8,
xlabel = {$x_1$ position (m)},
ylabel = {$x_2$ position (m)},
ylabel shift = -2pt,
every axis y label/.style={at={(ticklabel cs:0.5)},rotate=90,anchor=near ticklabel},
]

\addplot[color = blue,solid,thin,line join=round,unbounded coords=jump] table[x = x, y = y] {data.txt};

\end{axis}

\begin{axis}[%
height=\dataWidth,
width =\dataWidth,
at={(\dataWidth+4.4mm,0cm)},
scale only axis,
separate axis lines,
tick label style={/pgf/number format/fixed},
axis equal image=true,
yticklabels = {,,},
xtick distance=0.5,
xmin=-0.05,
xmax=1.55,
ymin=-0.8,
ymax=0.8,
xlabel = {$x_1$ position (m)},
every axis y label/.style={at={(ticklabel cs:0.5)},rotate=90,anchor=near ticklabel},
]

\addplot[color = blue,solid,line width = 0.5,line join=round,unbounded coords=jump,opacity=1.0] table[x = x, y = y] {data2.txt};
\coordinate (diana) at (axis cs:1.55,0); 
\end{axis}

\node[anchor=west,inner xsep = 0, inner ysep = 0, xshift=4.4mm] (image) at (diana){\includegraphics[height=\dataWidth]{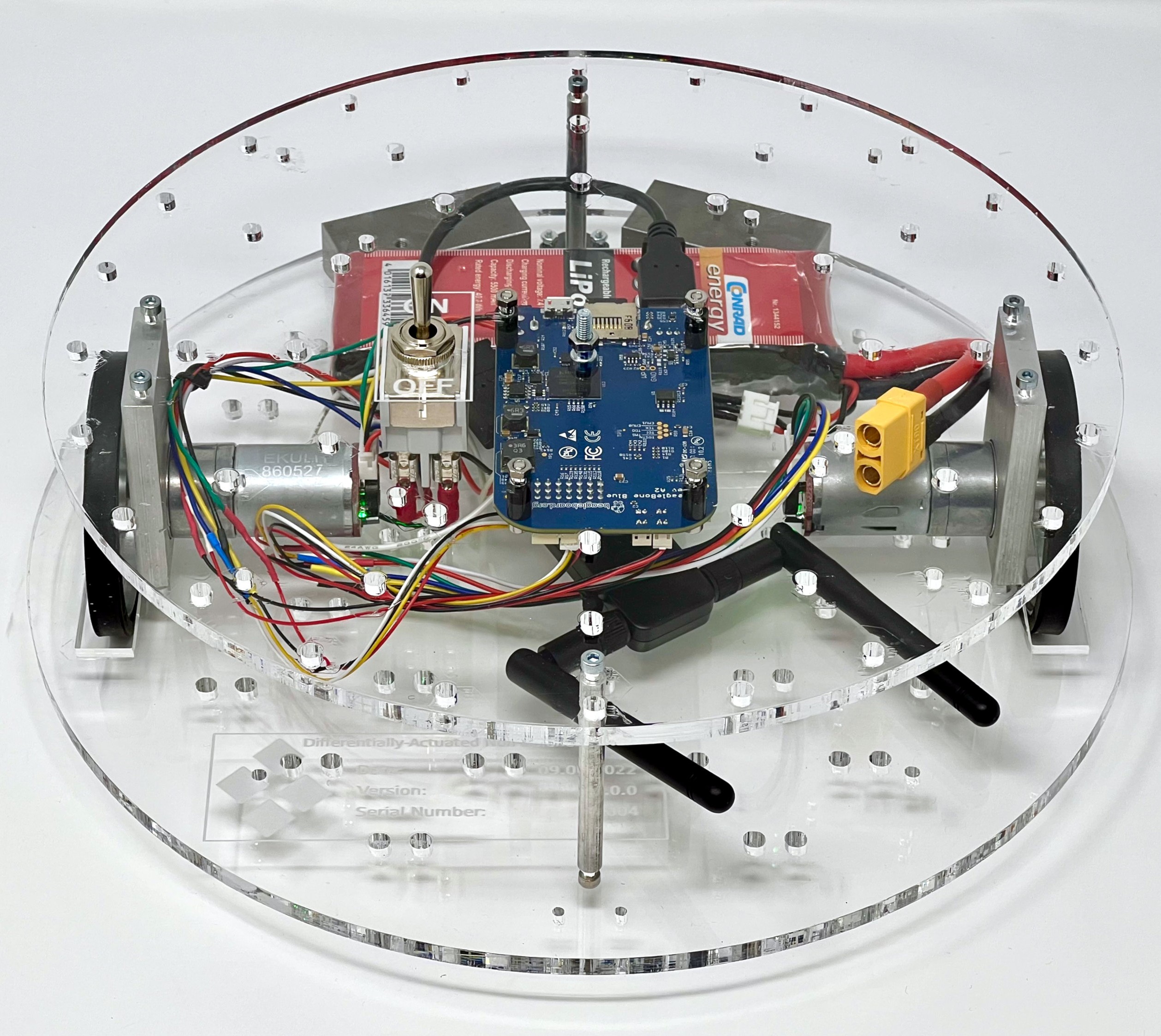}};
\end{tikzpicture}%

%% file: figures/tikz/mean_max_8.tex
\begin{tikzpicture}
\hspace{4pt}
\begin{axis}[
width = .405\textwidth,
height = \heightRealDataError,
at={(0.0, 0)},
legend cell align={left},
legend columns = 2,
legend style={
  fill opacity=1,
  draw opacity=1,
  text opacity=1,
  at={(1,1)},
  anchor=north east,
  column sep = 0.15cm
},
xlabel = {time (s)}, 
ylabel = {error},
xmin=-0, xmax=25,
ymin=-1e-05, ymax=0.005,
xshift=0.19cm
]
\addplot [line width = \linewidthErrorC, color = MaxError]
table {%
0 0
0.1 0.0218111330831636
0.2 0.0166472248805543
0.3 0.00237514206843605
0.4 0.00182170178453521
0.5 0.00164579073236717
0.6 0.00196000337950324
0.7 0.00113874255138009
0.8 0.00118168943661031
0.9 0.001160172931261
1 0.00122910121785218
1.1 0.00111081766217359
1.2 0.000858986488234555
1.3 0.00110773844630138
1.4 0.00119802088284138
1.5 0.00116935190467242
1.6 0.0019776421881197
1.7 0.00188687710664118
1.8 0.00093771466741924
1.9 0.00112615450499511
2 0.0011780655172553
2.1 0.00179470409992574
2.2 0.00147525362428379
2.3 0.00139894744667057
2.4 0.000819139504006065
2.5 0.00120920414319461
2.6 0.00140535390933018
2.7 0.00199599238447611
2.8 0.00185120341359877
2.9 0.000862463783887557
3 0.000601878517427685
3.1 0.000937473825799147
3.2 0.000696142325656951
3.3 0.00154049779804585
3.4 0.00150571912218871
3.5 0.000634425814140833
3.6 0.000667721772331938
3.7 0.000744632704745629
3.8 0.000615193939489468
3.9 0.00082220426978209
4 0.000848794282432759
4.1 0.00114440624315436
4.2 0.000698643711936882
4.3 0.000772726978441963
4.4 0.000664073251126432
4.5 0.000703441422968412
4.6 0.000918802170107213
4.7 0.000877287668899846
4.8 0.000692255669268265
4.9 0.000877904909881829
5 0.000555316114073289
5.1 0.000817334853110838
5.2 0.000835604530605776
5.3 0.0010664845591555
5.4 0.00106355815259028
5.5 0.000810109319470303
5.6 0.00103923566666399
5.7 0.000710252079863388
5.8 0.000673651708606017
5.9 0.00077613827449373
6 0.000808478237266405
6.1 0.000771293014645576
6.2 0.000820404289902603
6.3 0.000614788047022066
6.4 0.000777718111615948
6.5 0.000759756919552243
6.6 0.000628088093348934
6.7 0.000728338126275741
6.8 0.000693358227370197
6.9 0.000749205417638388
7 0.00118091528993192
7.1 0.0011590752213565
7.2 0.00115185915938922
7.3 0.000819431266890017
7.4 0.00110154293992482
7.5 0.00102460405937304
7.6 0.000799398114850884
7.7 0.000760792172164481
7.8 0.000846896035761227
7.9 0.000742296489287405
8 0.00122279941114923
8.1 0.00137293261394708
8.2 0.000810226480900624
8.3 0.000697674976649158
8.4 0.000745288842816009
8.5 0.00119846841920537
8.6 0.000740240646524749
8.7 0.000772921588417582
8.8 0.000714659783497115
8.9 0.00099181165354226
9 0.000739289932915314
9.1 0.000692945989677082
9.2 0.000711016777556627
9.3 0.00149795434969083
9.4 0.00139886349669039
9.5 0.000822840561000893
9.6 0.000744540838832495
9.7 0.000847148301396
9.8 0.000912370398483527
9.9 0.000746554300264008
10 0.00132921496233553
10.1 0.00148171118167327
10.2 0.000785310296790808
10.3 0.000809779553151078
10.4 0.000770628462325511
10.5 0.000910059184315224
10.6 0.000778920553660853
10.7 0.00184995117223489
10.8 0.0012298482705877
10.9 0.00119022435436737
11 0.00164299566278309
11.1 0.00202733599503063
11.2 0.000838711797071759
11.3 0.00217386104988614
11.4 0.0012333802140042
11.5 0.00189238549807984
11.6 0.0010879664575399
11.7 0.00230606401196725
11.8 0.0015766531716285
11.9 0.000618100910581359
12 0.00121613108246024
12.1 0.00128013018491497
12.2 0.00110259818799942
12.3 0.00130508326151106
12.4 0.000875945817304124
12.5 0.00149161012416394
12.6 0.000845522104393796
12.7 0.00144503109514916
12.8 0.000878380558363015
12.9 0.00111058310280176
13 0.00125020365462378
13.1 0.00117653798182924
13.2 0.00167261761307526
13.3 0.000706254580971073
13.4 0.00124848406152861
13.5 0.00100861398260138
13.6 0.00129079090248657
13.7 0.00171234480910751
13.8 0.00175539592106978
13.9 0.000890539130872647
14 0.00113268966958203
14.1 0.00104536475711799
14.2 0.000839373409885747
14.3 0.000855888971494867
14.4 0.000860975958597921
14.5 0.000962882131407204
14.6 0.000967954458361552
14.7 0.00110905231084328
14.8 0.00106372424521503
14.9 0.000826308320354236
15 0.00101990266120665
15.1 0.00115604399624758
15.2 0.00164676320413975
15.3 0.000929545497489082
15.4 0.00116936951785769
15.5 0.00124384607253603
15.6 0.00171070392573958
15.7 0.00106765678032582
15.8 0.00118231574128314
15.9 0.00141131087891961
16 0.000680310797916312
16.1 0.0012520256879146
16.2 0.00104163868575545
16.3 0.000654455363419246
16.4 0.00103432289092331
16.5 0.000818533939149639
16.6 0.0010126848606802
16.7 0.000540852645001082
16.8 0.00112483285466042
16.9 0.000561379653397216
17 0.000863098135906454
17.1 0.00100814126578749
17.2 0.000792091094689208
17.3 0.000801836501969912
17.4 0.000756639942120807
17.5 0.00101485717720009
17.6 0.00131399204623153
17.7 0.000834961685914142
17.8 0.0008664539901077
17.9 0.000701567022655454
18 0.000684086756828024
18.1 0.000823827113785157
18.2 0.000923066866633415
18.3 0.00114420130416271
18.4 0.000974713180786147
18.5 0.000975679347363912
18.6 0.000850070834138813
18.7 0.000836566984504858
18.8 0.000955138947456466
18.9 0.000917611817943651
19 0.00108634755477606
19.1 0.00112730293463918
19.2 0.00108351929691235
19.3 0.00117049437419037
19.4 0.000807565662042005
19.5 0.000873083218083089
19.6 0.000865410003871241
19.7 0.000913135349093921
19.8 0.000760760521778913
19.9 0.000935465583664372
20 0.000820498552890802
20.1 0.000917520844378701
20.2 0.000771878891951089
20.3 0.000997368640354455
20.4 0.00184888836018966
20.5 0.00166946068983158
20.6 0.00112651446631485
20.7 0.00107210024386617
20.8 0.000915537878561935
20.9 0.000910702761685879
21 0.00115334968575375
21.1 0.000890204377010191
21.2 0.000804008786428191
21.3 0.00106454290202765
21.4 0.000795773304781374
21.5 0.00125189571371278
21.6 0.00104055052614912
21.7 0.00108006238185511
21.8 0.000910216949063225
21.9 0.000955947709258221
22 0.00130232038045185
22.1 0.000648755411043843
22.2 0.00136544416518835
22.3 0.00097494877579632
22.4 0.000582040725109847
22.5 0.000910709067860922
22.6 0.000738551009741783
22.7 0.000998407269776071
22.8 0.000969878703478162
22.9 0.00168146758319692
23 0.000751994745763455
23.1 0.00100768668364692
23.2 0.00102413963640494
23.3 0.00083405054556639
23.4 0.00134840860550717
23.5 0.00117395378508981
23.6 0.000935592207059303
23.7 0.000839910285277137
23.8 0.00214663118502847
23.9 0.00219298516068275
24 0.00232767289758298
24.1 0.00149881087644298
24.2 0.00104957540397154
24.3 0.00221402854508365
24.4 0.00124645959062284
24.5 0.000953242754624308
24.6 0.00229528977055165
24.7 0.00222197648060372
24.8 0.00121233139270325
};\label{plot:MaxErrorEight}
\addlegendentry{$e_{\text{max}}$}
\addplot [line width = \linewidthErrorC, color = AvgErr]
table {%
0 0
0.1 0.0178480138587105
0.2 0.00346136960533238
0.3 0.00168251853907658
0.4 0.00112301906284021
0.5 0.000832727710936349
0.6 0.000510055158639128
0.7 0.00039842563906734
0.8 0.000393303143266708
0.9 0.000459975456877203
1 0.000406231109289026
1.1 0.000555252553064137
1.2 0.000430449743368592
1.3 0.000508456284717768
1.4 0.000515604725540065
1.5 0.000412531841441262
1.6 0.000548886118901324
1.7 0.000572254182847836
1.8 0.000413668561277504
1.9 0.000350319824151475
2 0.000466797808874442
2.1 0.000508889721853971
2.2 0.000500003219953263
2.3 0.000445229705934446
2.4 0.000310428674926053
2.5 0.000394313043558444
2.6 0.000406539918632542
2.7 0.000460363173280814
2.8 0.000448415712219654
2.9 0.000384005029749448
3 0.000316303124456396
3.1 0.000425818101627574
3.2 0.000371456514660979
3.3 0.000462551244798035
3.4 0.000436692489068627
3.5 0.000381163138117614
3.6 0.00039176839023812
3.7 0.000406415030087471
3.8 0.00037487122958732
3.9 0.000501486518207436
4 0.000410233138983916
4.1 0.000520552759753522
4.2 0.00040432676457731
4.3 0.000469595926971492
4.4 0.000388266973502595
4.5 0.000410571129221132
4.6 0.000493032029914097
4.7 0.000369370421270014
4.8 0.000472103466127895
4.9 0.000472659554082058
5 0.000402803669194487
5.1 0.000420398327752655
5.2 0.000381906627062597
5.3 0.000489934800507741
5.4 0.000484875759501379
5.5 0.000350414149606082
5.6 0.000460121261847649
5.7 0.000348656066347556
5.8 0.000289323376865398
5.9 0.000409936315273775
6 0.00046787049413993
6.1 0.000369304919060326
6.2 0.000446892262918518
6.3 0.000407756887910023
6.4 0.000449127108704602
6.5 0.000341993446420941
6.6 0.000389066866927044
6.7 0.000395046575115106
6.8 0.000356410059560883
6.9 0.00042916620920751
7 0.000526995891733802
7.1 0.000431426161140984
7.2 0.000429585333430335
7.3 0.000502138091763193
7.4 0.000478222509314336
7.5 0.000559873957220904
7.6 0.000438081908879151
7.7 0.00043099523650966
7.8 0.000485906921630406
7.9 0.000448709840090899
8 0.000456239863236378
8.1 0.000546282285420562
8.2 0.000342275242033338
8.3 0.000469713719297109
8.4 0.000441286446444214
8.5 0.000495415053529066
8.6 0.000446450024972111
8.7 0.000395001223000835
8.8 0.000459130673366819
8.9 0.000426615865692915
9 0.000407747417659968
9.1 0.000429676976032896
9.2 0.000352723823080096
9.3 0.000472008980913212
9.4 0.000513556058988567
9.5 0.00039848666886367
9.6 0.000378136393244121
9.7 0.000363628479878839
9.8 0.000368575564416556
9.9 0.000350282378496879
10 0.000486508656354637
10.1 0.000387293499335707
10.2 0.000369794272236367
10.3 0.000393579468817532
10.4 0.000386048597799702
10.5 0.000363003541448739
10.6 0.000352980111642184
10.7 0.000477367395163282
10.8 0.000394622629634395
10.9 0.000364558190993254
11 0.000409089240999483
11.1 0.000472634520783616
11.2 0.000270291584728241
11.3 0.000455967871704817
11.4 0.000363674070763994
11.5 0.000470122167192538
11.6 0.000348864229052387
11.7 0.000549366138935868
11.8 0.000409974164874536
11.9 0.000325056663457393
12 0.000466858269415193
12.1 0.000558189816159613
12.2 0.000369011016748082
12.3 0.00049120527176261
12.4 0.000412259493327199
12.5 0.000511784845452854
12.6 0.000418524372737589
12.7 0.000445136250651688
12.8 0.000432996129376773
12.9 0.000379596788669839
13 0.000444876207956507
13.1 0.000415344756153192
13.2 0.000487818652709673
13.3 0.000342253485153672
13.4 0.000543839373042666
13.5 0.000418155145669528
13.6 0.000447382943225051
13.7 0.000398282718790385
13.8 0.000482306053659726
13.9 0.000387381493957682
14 0.000362125104589586
14.1 0.000416322086396805
14.2 0.000423200480849395
14.3 0.000479641963938867
14.4 0.000376648615696125
14.5 0.000453031813520734
14.6 0.000439166948181694
14.7 0.000452266151005869
14.8 0.000524598956297956
14.9 0.000405980821282593
15 0.000431857558438903
15.1 0.000511277443467781
15.2 0.000477582582202853
15.3 0.000388135186018543
15.4 0.000401566719309941
15.5 0.000446522271405705
15.6 0.000503569354327938
15.7 0.000346293264076215
15.8 0.000400164149454075
15.9 0.000457355318216753
16 0.000351460846695983
16.1 0.000416329965784737
16.2 0.000307078470562202
16.3 0.000337597155558062
16.4 0.000423300138727769
16.5 0.000274128730662479
16.6 0.000372023919929724
16.7 0.000274101619856441
16.8 0.000351328322707309
16.9 0.000300200108256463
17 0.000329771789689699
17.1 0.000329839822688807
17.2 0.000330350917527844
17.3 0.000294142621253281
17.4 0.000373085039688761
17.5 0.000359044034294392
17.6 0.000391947828394907
17.7 0.000414418912398849
17.8 0.000303851429429684
17.9 0.000346581854604036
18 0.00031640577176365
18.1 0.000303341624172758
18.2 0.000401556537244306
18.3 0.000423566518241002
18.4 0.000464025416571
18.5 0.000407249068921094
18.6 0.000320415946483372
18.7 0.000289999772282618
18.8 0.00038411575699688
18.9 0.000356588302546137
19 0.000370467764026846
19.1 0.000420119793670597
19.2 0.000386689379562959
19.3 0.000422341091129088
19.4 0.000291526663758252
19.5 0.000271109287898287
19.6 0.000253238999638801
19.7 0.000298875123203227
19.8 0.000347245253199604
19.9 0.000322733729003605
20 0.000329835995168627
20.1 0.000290223530729577
20.2 0.000335551201198197
20.3 0.000376481898032925
20.4 0.00046458626980805
20.5 0.000484482708865416
20.6 0.000402459820004954
20.7 0.000445126201743358
20.8 0.000461805980666489
20.9 0.000373072512368866
21 0.000467752076588497
21.1 0.000430563844468952
21.2 0.000435266356434795
21.3 0.00040836033747623
21.4 0.000428583503506233
21.5 0.000472270633043583
21.6 0.000401556867256419
21.7 0.000382620727688219
21.8 0.000424168736243815
21.9 0.000428736733461563
22 0.000410154826258137
22.1 0.000344937418226348
22.2 0.000398575762192661
22.3 0.000382415433816121
22.4 0.000348613270782797
22.5 0.00032815032326095
22.6 0.000361925036810857
22.7 0.000374005753111173
22.8 0.000376402951282942
22.9 0.000374583268279
23 0.000325324934572279
23.1 0.00039975599296835
23.2 0.000469548810699636
23.3 0.000442858059960306
23.4 0.000414090548263624
23.5 0.000390415846222797
23.6 0.00039288643650547
23.7 0.000451894185899233
23.8 0.000543890945595062
23.9 0.000658471065129718
24 0.000614187978679585
24.1 0.000513351091487459
24.2 0.000463115324423361
24.3 0.000632837339420454
24.4 0.000531969845428524
24.5 0.000589684477361628
24.6 0.000605776008354775
24.7 0.000628512252509024
24.8 0.000434691345317759
};\label{plot:AvgErrorEight}
\addlegendentry{$e_{\text{avg}}$}
\addplot [line width = \linewidthErrorStdVar, color = Koopman] 
table {%
0 0
0.1 0.0190491623042049
0.2 0.00704799793596951
0.3 0.00209433207062622
0.4 0.00141757715673368
0.5 0.00126606084347488
0.6 0.000960984675418757
0.7 0.000641400990496132
0.8 0.00066883080750079
0.9 0.000717208566771342
1 0.000726985037606839
1.1 0.000798676786369337
1.2 0.000653141302798547
1.3 0.000756874725582265
1.4 0.000759747238749241
1.5 0.000659551314029731
1.6 0.00100469887992618
1.7 0.00104661852248414
1.8 0.000634745004462205
1.9 0.000601637456269032
2 0.000744077690348807
2.1 0.000936294595837474
2.2 0.000911664389946073
2.3 0.000779135713400769
2.4 0.000549072589617891
2.5 0.000686711600912299
2.6 0.000720681459784784
2.7 0.000949195029965004
2.8 0.000845793796689429
2.9 0.00059050598788353
3 0.000479733914386366
3.1 0.000689694619494556
3.2 0.00052726287327279
3.3 0.000823669505887278
3.4 0.000746468967988175
3.5 0.000522660558264304
3.6 0.000596470542404951
3.7 0.00057935387432754
3.8 0.000552009051272775
3.9 0.000671367465401712
4 0.000643582748701368
4.1 0.000785011477059047
4.2 0.000560775512567449
4.3 0.000613629363907817
4.4 0.000539554672064954
4.5 0.00059997072697477
4.6 0.000736118025849699
4.7 0.000555668625347132
4.8 0.000643223978540583
4.9 0.000645120084680304
5 0.000500573044597864
5.1 0.000624814278455164
5.2 0.000602188841672137
5.3 0.000726969353428992
5.4 0.000732803060714139
5.5 0.000526786313767674
5.6 0.000686239836265107
5.7 0.000526426780774891
5.8 0.00046666234411077
5.9 0.000559075909999121
6 0.00061391540606144
6.1 0.000510359012800975
6.2 0.000649285063093051
6.3 0.000548234602110887
6.4 0.000645037651629012
6.5 0.000543136218786295
6.6 0.000503403229479756
6.7 0.000533273733853097
6.8 0.000534230701530469
6.9 0.000619267267535731
7 0.000743757749824733
7.1 0.000675381957968618
7.2 0.000646276695772685
7.3 0.000663292072722796
7.4 0.000694274744761798
7.5 0.000794629969010058
7.6 0.000613799294867365
7.7 0.000640852096864321
7.8 0.00066547277057286
7.9 0.000611785820290137
8 0.000720057143023211
8.1 0.000816686704953236
8.2 0.000508799338256045
8.3 0.000607513023361248
8.4 0.000612291875276559
8.5 0.000775763671470604
8.6 0.000612188971036919
8.7 0.000553563807920814
8.8 0.000621905387896033
8.9 0.000659109432182751
9 0.000551601839298667
9.1 0.000620893652175312
9.2 0.000497737426112242
9.3 0.000766705295092424
9.4 0.00080693001109161
9.5 0.000601963945476729
9.6 0.000543125769688944
9.7 0.000589988772867731
9.8 0.000553228935050132
9.9 0.000557907471136746
10 0.000831444669174632
10.1 0.000740643326633164
10.2 0.000526815249595294
10.3 0.000598604560258954
10.4 0.000567481981642056
10.5 0.000586054201195979
10.6 0.000518233360285903
10.7 0.000890074981821544
10.8 0.000673642318372208
10.9 0.000650398447281175
11 0.000792749094923946
11.1 0.000977880371079086
11.2 0.000461879857413873
11.3 0.00100417593850639
11.4 0.000616496389142681
11.5 0.000929134098199974
11.6 0.00061355858435514
11.7 0.00105814811647695
11.8 0.000781654144705779
11.9 0.000491279108492736
12 0.000747485905768025
12.1 0.000822585800409748
12.2 0.000621448204458177
12.3 0.000769011509197561
12.4 0.000647782961953572
12.5 0.000842319601431871
12.6 0.000623350923600542
12.7 0.000746533840858154
12.8 0.000654978789557511
12.9 0.000634524700572168
13 0.000701602799976788
13.1 0.000681762106502567
13.2 0.000832581537980088
13.3 0.000541192616943161
13.4 0.000837912469927652
13.5 0.000653210949514678
13.6 0.000746920979754525
13.7 0.000839372569722776
13.8 0.000861769412222676
13.9 0.000591991100559362
14 0.000626435891966447
14.1 0.000646156553307237
14.2 0.000622022785708915
14.3 0.000649987545280624
14.4 0.000551577450550621
14.5 0.000677100499293426
14.6 0.00068094352182683
14.7 0.00068606814308832
14.8 0.000722305566344855
14.9 0.000610689440812786
15 0.00064545596406583
15.1 0.000832824889111897
15.2 0.000863432543917598
15.3 0.000607931148874746
15.4 0.000681425773467903
15.5 0.000736693843302295
15.6 0.000847485238137124
15.7 0.00060706670052829
15.8 0.000726666129415896
15.9 0.000808291342990077
16 0.000502339513068011
16.1 0.000668091195694862
16.2 0.00055875525650616
16.3 0.000501236295340802
16.4 0.000678281152339475
16.5 0.000457789636647539
16.6 0.00061942054517356
16.7 0.000438487568638623
16.8 0.00060783193890024
16.9 0.00045381109571069
17 0.000517924223929636
17.1 0.000549043180377354
17.2 0.000506996848825398
17.3 0.000496346673167572
17.4 0.000568318301756044
17.5 0.00063807028379851
17.6 0.000692705097697475
17.7 0.000622927400017558
17.8 0.000540944164608012
17.9 0.0005320688714957
18 0.000497834412193976
18.1 0.000476286042149753
18.2 0.000640553296915761
18.3 0.000746550881252391
18.4 0.000741187358285568
18.5 0.000629851435742356
18.6 0.000539604046365171
18.7 0.00051569523079599
18.8 0.000615845293594637
18.9 0.000609490255031382
19 0.000660391458538922
19.1 0.000694009033884897
19.2 0.000728441375060119
19.3 0.000689015111416167
19.4 0.000463447465891365
19.5 0.000469330004903184
19.6 0.000464847490359082
19.7 0.000504010071642718
19.8 0.000545738944002571
19.9 0.000540012333695694
20 0.000532721388344323
20.1 0.000508567248209636
20.2 0.000493810946005179
20.3 0.000602200887856431
20.4 0.000876491982517506
20.5 0.000856026577168347
20.6 0.000675643954390011
20.7 0.000664216059162964
20.8 0.000665924981696889
20.9 0.000575347268177347
21 0.000690067797652509
21.1 0.000621181605321642
21.2 0.000609344606379007
21.3 0.00063491509839819
21.4 0.00059085629890106
21.5 0.000720487744503473
21.6 0.000617710305254994
21.7 0.000624787291581173
21.8 0.000695000704547913
21.9 0.000657199970251283
22 0.000702514541826663
22.1 0.000514260290490728
22.2 0.000733410729314145
22.3 0.000634392992172463
22.4 0.000524889539428506
22.5 0.000521319453181333
22.6 0.000566973256150851
22.7 0.000602670181775924
22.8 0.000648425372702121
22.9 0.000749539161431323
23 0.000507764692119787
23.1 0.000621377325213472
23.2 0.000741372481497649
23.3 0.000681557697444227
23.4 0.000750424692364996
23.5 0.000647469246695626
23.6 0.000630187916473506
23.7 0.000647260038736539
23.8 0.00105240629026901
23.9 0.00115693686448696
24 0.00116904977700385
24.1 0.000830646641539638
24.2 0.000731951067800125
24.3 0.0011105309084948
24.4 0.00077699742736547
24.5 0.000834306068700037
24.6 0.00113043197927471
24.7 0.00114667297698243
24.8 0.0007687305639908
};\label{plot:hErrorEight}
\addlegendentry{$\sigma$}
\end{axis}
\begin{axis}[
width = .405\textwidth,
height = \heightRealDataError,
at={(.354\textwidth, 0)},
legend columns = 1,
legend style={
  fill opacity=1,
  draw opacity=1,
  text opacity=1,
  at={(1,1.1)},
  anchor=north east,
  column sep = 0.2cm
},
xlabel = {time (s)}, 
xmin=-0, xmax=28,
ymin=-0.000317420907404805, ymax=0.00666583905550091,
]
\addplot [line width = \linewidthErrorC, color = MaxError]
table {%
0 0
0.1 0.00243030151026045
0.2 0.00191673452960251
0.3 0.000887512761918704
0.4 0.00147929005628157
0.5 0.00173792875028588
0.6 0.000984666627024665
0.7 0.00138936251650066
0.8 0.00229776125775171
0.9 0.00124260889840826
1 0.00189845088152088
1.1 0.00101863278442778
1.2 0.00224207344283344
1.3 0.0022258720196888
1.4 0.0021716530098494
1.5 0.00188511936984768
1.6 0.00193513933205989
1.7 0.00179741673935735
1.8 0.000593499900620881
1.9 0.000710950602873612
2 0.000859365209289473
2.1 0.0020469892391188
2.2 0.000764688589694938
2.3 0.00214234926890111
2.4 0.00189678924696314
2.5 0.000563620497323481
2.6 0.00219978509915878
2.7 0.00221348212684458
2.8 0.000810499743372499
2.9 0.00225232865072449
3 0.000809805997244321
3.1 0.00178150792933667
3.2 0.00192789585074392
3.3 0.000782017002823046
3.4 0.0019620956325225
3.5 0.000824355191515104
3.6 0.00247622572403226
3.7 0.00185794230652155
3.8 0.00178629763675698
3.9 0.00214665061286236
4 0.00231085572237006
4.1 0.000807539876157421
4.2 0.00101025433243081
4.3 0.00243985277031156
4.4 0.00267844233203417
4.5 0.000632158133444315
4.6 0.00236179027461731
4.7 0.00216664205742035
4.8 0.00229167212816204
4.9 0.000887367341716449
5 0.000816247027397059
5.1 0.00180220105999559
5.2 0.00143405782346682
5.3 0.00114237921381472
5.4 0.0016905605643704
5.5 0.000872297501729055
5.6 0.000786497432161249
5.7 0.000923499840409351
5.8 0.000943175467467437
5.9 0.000768974558825933
6 0.00139514376858888
6.1 0.00080717308555432
6.2 0.00115309380050881
6.3 0.00126300518599934
6.4 0.00136787680624442
6.5 0.00151434580993336
6.6 0.00179654804122978
6.7 0.00187071119531102
6.8 0.00147605557992692
6.9 0.00121232198022002
7 0.000992209322900209
7.1 0.000727030637053469
7.2 0.000532347953306163
7.3 0.00089391807343616
7.4 0.000511369134792497
7.5 0.002523359497698
7.6 0.00210656671591806
7.7 0.000597782799362139
7.8 0.0011630886536476
7.9 0.00177604083530233
8 0.000980652255753125
8.1 0.00200732865558027
8.2 0.00225753547755002
8.3 0.00132908370495918
8.4 0.00185179998061135
8.5 0.0014528064799741
8.6 0.00203479128465893
8.7 0.000726661022930454
8.8 0.000683253822841503
8.9 0.00193014234571764
9 0.00202448586206326
9.1 0.00100440579503102
9.2 0.00180037667394974
9.3 0.00167056279268262
9.4 0.00169140486977675
9.5 0.00212435172756484
9.6 0.00164306240893193
9.7 0.00198166297198967
9.8 0.00215660553635352
9.9 0.00103633750949908
10 0.00213614381702186
10.1 0.00277807559546614
10.2 0.00260630113932305
10.3 0.00271626359823287
10.4 0.00200373018716632
10.5 0.00210952693212198
10.6 0.00238031225727075
10.7 0.00191676998334233
10.8 0.00196454227784894
10.9 0.00209056192580351
11 0.00213089736853111
11.1 0.00171399170471313
11.2 0.000779955871224051
11.3 0.00198517913518675
11.4 0.00159932840592942
11.5 0.00204663745300815
11.6 0.00108185929836322
11.7 0.0025654797646767
11.8 0.00244275181842232
11.9 0.0010207147104581
12 0.000873297947506
12.1 0.0010207906807331
12.2 0.00186393047835346
12.3 0.00183776738581444
12.4 0.00214156724563992
12.5 0.00303388734482888
12.6 0.00396860868117939
12.7 0.00435221970923234
12.8 0.00385769220572877
12.9 0.0063484181480961
13 0.00460070297429933
13.1 0.00192794149846424
13.2 0.00383024568133048
13.3 0.00318734336005937
13.4 0.00175081086988246
13.5 0.00152045307790448
13.6 0.00145744674204857
13.7 0.00168174721564103
13.8 0.00186233607346212
13.9 0.00200740716275265
14 0.00208275421749735
14.1 0.00217996866459722
14.2 0.00157380481020825
14.3 0.00158492694011368
14.4 0.00123631953629882
14.5 0.000960540180434862
14.6 0.000900047915273402
14.7 0.000830389347365177
14.8 0.000437868869268062
14.9 0.00265153735603369
15 0.00398114002134728
15.1 0.00162419091478261
15.2 0.00278662117386217
15.3 0.00485917900057668
15.4 0.0037783147545486
15.5 0.00182475866674973
15.6 0.00232995110054523
15.7 0.00254695445922885
15.8 0.00145704664756429
15.9 0.002241757458959
16 0.00278628900598818
16.1 0.00113262940424731
16.2 0.0009194331613674
16.3 0.00108647130571415
16.4 0.000887106629106828
16.5 0.00203407739287681
16.6 0.000713281947606964
16.7 0.00253301138017987
16.8 0.000982420423433222
16.9 0.00175504518469364
17 0.0014870533802038
17.1 0.000811596726628912
17.2 0.000713671015137884
17.3 0.000999677246703432
17.4 0.000722599665164494
17.5 0.000709777821357932
17.6 0.00207483829897343
17.7 0.000758448776913675
17.8 0.00227394507767697
17.9 0.00216715332566875
18 0.00191911220194457
18.1 0.0021065810019956
18.2 0.00244716642800937
18.3 0.000856227162847044
18.4 0.000725173073547468
18.5 0.000964527619822043
18.6 0.00186314666903763
18.7 0.00205928450844084
18.8 0.00192272024685532
18.9 0.00169332589346587
19 0.000735786607242953
19.1 0.000851677116652567
19.2 0.000921440732309084
19.3 0.00190681945953216
19.4 0.00072744868301746
19.5 0.0020659866614276
19.6 0.00169946990664866
19.7 0.00260046455878947
19.8 0.000967945926301732
19.9 0.00279365511543098
20 0.00549225916542742
20.1 0.00451843718541041
20.2 0.00224661906006302
20.3 0.00504711102305296
20.4 0.00417957480572267
20.5 0.00116080372431168
20.6 0.00135190412730632
20.7 0.00276397813032723
20.8 0.00248798923109946
20.9 0.000932350012909134
21 0.000957704745632271
21.1 0.00130583791146355
21.2 0.00124959969795433
21.3 0.00165879907310727
21.4 0.00176299017569233
21.5 0.00181362965863251
21.6 0.00138068246829359
21.7 0.00128304359004505
21.8 0.000986483816859094
21.9 0.000694766090812802
22 0.000564037451134075
22.1 0.000368499422788179
22.2 0.00036194784841371
22.3 0.00278442971441178
22.4 0.00336578500244648
22.5 0.00509657721859998
22.6 0.00375243135686919
22.7 0.00555778335433584
22.8 0.00329322626027611
22.9 0.00219074291230735
23 0.00301357022174283
23.1 0.00143931114164639
23.2 0.000931945075180504
23.3 0.0023225871822788
23.4 0.00188611669195171
23.5 0.000740280745198771
23.6 0.00177053954195161
23.7 0.00174170258962898
23.8 0.00207945674732425
23.9 0.00259061309482629
24 0.00364455038119097
24.1 0.00309620782487069
24.2 0.00183604988116839
24.3 0.00124248464761039
24.4 0.00118280863751591
24.5 0.00173520511362815
24.6 0.00234706845673169
24.7 0.00159751522145684
24.8 0.00167149590959325
24.9 0.00126164160975007
25 0.0017374205979871
25.1 0.00132544310008675
25.2 0.00330046265004049
25.3 0.00115380073160819
25.4 0.00125602088620519
25.5 0.00236295226741401
25.6 0.00130648748224772
25.7 0.00220262993872144
25.8 0.00265942026206182
25.9 0.00211014041459084
26 0.00122530593807025
26.1 0.0021134944749626
26.2 0.00188353321088364
26.3 0.000884879942476731
26.4 0.0008965119956245
26.5 0.000922555471395546
26.6 0.00214286757233242
26.7 0.00092153225345721
26.8 0.00191558358185165
26.9 0.00215299109539737
27 0.000811119688002781
27.1 0.000970815985005502
27.2 0.00138913004191141
27.3 0.00292941649705783
27.4 0.00362685701682557
27.5 0.00542868401636261
27.6 0.00549505358274129
27.7 0.00426660418179206
27.8 0.00095564338549811
27.9 0.00479671384427148
};
\addplot [line width = \linewidthErrorC, color = AvgErr]
table {%
0 0
0.1 0.00214995191340496
0.2 0.00140390878492252
0.3 0.000325771334712744
0.4 0.000707136938050266
0.5 0.000686346920103303
0.6 0.000612988560202573
0.7 0.000583400299982956
0.8 0.000719823939759715
0.9 0.000864997931102958
1 0.00052815281364469
1.1 0.000538933338336908
1.2 0.000571053669130864
1.3 0.000479849296754965
1.4 0.000451481135906668
1.5 0.000472910954232216
1.6 0.000427455983453428
1.7 0.000625226386542763
1.8 0.000337441923252853
1.9 0.00038065001456982
2 0.000395764417683126
2.1 0.000464963769876949
2.2 0.00035266522871744
2.3 0.000464799198525325
2.4 0.000481319712855152
2.5 0.000313318408054501
2.6 0.000605097125228146
2.7 0.00061001529701726
2.8 0.000301494217292217
2.9 0.000478250630457454
3 0.000392712016535797
3.1 0.000395206497805362
3.2 0.000455046590051422
3.3 0.00038966441653259
3.4 0.000409261791102267
3.5 0.000344181665219991
3.6 0.000635458756570036
3.7 0.00063378832933252
3.8 0.000469756944061987
3.9 0.000437539783843503
4 0.000521003711479189
4.1 0.00044677503997516
4.2 0.000456797602215318
4.3 0.000518227294795556
4.4 0.000565740307489386
4.5 0.000327625306384091
4.6 0.000536159850105096
4.7 0.000486118642494934
4.8 0.000650935366937868
4.9 0.000353064760505279
5 0.000373711609313808
5.1 0.00130586769924227
5.2 0.000435607652595198
5.3 0.000737649638800371
5.4 0.000684542273140462
5.5 0.000561942285627596
5.6 0.000486208606081046
5.7 0.000522777197744279
5.8 0.000624716647932447
5.9 0.000571820735528701
6 0.00112529799493495
6.1 0.000652499637793528
6.2 0.000856087441237497
6.3 0.000995388253914678
6.4 0.00112440181795449
6.5 0.00133042389068265
6.6 0.0015191581409735
6.7 0.001571226656355
6.8 0.00118946739823819
6.9 0.00105206680055683
7 0.000835633079482644
7.1 0.000611691987573463
7.2 0.000431137244228736
7.3 0.000266044857639398
7.4 0.000195703850793922
7.5 0.00212078048774427
7.6 0.00155642261956804
7.7 0.00028489318032669
7.8 0.000687356006550015
7.9 0.00053559434407874
8 0.000664513043564771
8.1 0.000576843838990278
8.2 0.000695568203658888
8.3 0.000559727837042746
8.4 0.000566073788240878
8.5 0.000531366922874246
8.6 0.000516480345840153
8.7 0.000347040172453653
8.8 0.000365827678300548
8.9 0.000566542487376736
9 0.000508736585392571
9.1 0.000511174741753438
9.2 0.000516029254541366
9.3 0.000587911605012655
9.4 0.000450407186849406
9.5 0.000512026476999501
9.6 0.000520099723185061
9.7 0.000633628321411928
9.8 0.000560893066014673
9.9 0.000482168326467034
10 0.000564006607477693
10.1 0.000592411507606343
10.2 0.000584972828531718
10.3 0.000695769729959216
10.4 0.000527287440905794
10.5 0.000535528891005092
10.6 0.000665615724815263
10.7 0.000607125881588774
10.8 0.000541780533218091
10.9 0.000652285046095762
11 0.00058188119151537
11.1 0.000563154701898057
11.2 0.000419523060531403
11.3 0.000559617099971077
11.4 0.00056311691335904
11.5 0.000511718721579497
11.6 0.000470395280087649
11.7 0.000675736861901787
11.8 0.000716467016882296
11.9 0.000538437159376277
12 0.000455502673828448
12.1 0.000477535363149772
12.2 0.000643668064973019
12.3 0.00067624147241258
12.4 0.000548425057815169
12.5 0.00143661106995938
12.6 0.0010271696462158
12.7 0.00150554114747026
12.8 0.00123743288084334
12.9 0.00102404185217932
13 0.000992617555657845
13.1 0.000901318471839698
13.2 0.000867369894888715
13.3 0.000903443027006259
13.4 0.00102486851174129
13.5 0.000731886613897782
13.6 0.000843636407792331
13.7 0.00103808270038527
13.8 0.00129738101032276
13.9 0.00155455636741119
14 0.00158765744782914
14.1 0.00164914197219794
14.2 0.00125002434928474
14.3 0.00107764799194429
14.4 0.000877825884688674
14.5 0.000691004159938887
14.6 0.000469921142148943
14.7 0.000435227796879541
14.8 0.000189447317226511
14.9 0.00223967419259794
15 0.00146739993371378
15.1 0.00067336244993143
15.2 0.000885448347150335
15.3 0.000878303847451836
15.4 0.000899321704972797
15.5 0.000706762391558523
15.6 0.000968143705687901
15.7 0.000726866571646136
15.8 0.000565703507336729
15.9 0.000589880319663505
16 0.000577132008488073
16.1 0.000559187951263492
16.2 0.000497075761598006
16.3 0.000542686484735261
16.4 0.000368554820265425
16.5 0.000592639915394481
16.6 0.000406815337941095
16.7 0.000633421975816608
16.8 0.000553955311748541
16.9 0.000543830183296829
17 0.000527270447049267
17.1 0.000406570422506521
17.2 0.000394742442383238
17.3 0.000443165705924016
17.4 0.00033337753123386
17.5 0.000392973397729035
17.6 0.000498237758218976
17.7 0.000366655628030927
17.8 0.000483910914068369
17.9 0.000471149627714541
18 0.000493433784956355
18.1 0.000595360587991296
18.2 0.000634694089613901
18.3 0.000396120589897357
18.4 0.000349707483941994
18.5 0.000385564908649255
18.6 0.000471218520208343
18.7 0.000486233407519572
18.8 0.000417256346699489
18.9 0.000455494018989065
19 0.000390969839153773
19.1 0.000352950107890155
19.2 0.000363033863201353
19.3 0.000476729456512644
19.4 0.00034957101634229
19.5 0.000544746683701971
19.6 0.000576327251673644
19.7 0.000578303065939675
19.8 0.000486749382779984
19.9 0.00172176512305615
20 0.00107171717672001
20.1 0.00111396626057107
20.2 0.000862708812150197
20.3 0.00118331864325502
20.4 0.000919972921973535
20.5 0.000660063793320172
20.6 0.000801389524190682
20.7 0.000727312084813533
20.8 0.00115309898085549
20.9 0.000745892930231987
21 0.000838766433658393
21.1 0.00105064703348448
21.2 0.00115050115523215
21.3 0.00146319061685302
21.4 0.00161381918108652
21.5 0.00158208788964914
21.6 0.00123038367526442
21.7 0.00105650889448498
21.8 0.000842921694452999
21.9 0.000610425014588016
22 0.000427306693801245
22.1 0.000205696557186717
22.2 0.000109601770136235
22.3 0.00213836109943829
22.4 0.00173787250266421
22.5 0.000686603272001525
22.6 0.00101524112456659
22.7 0.0012008115897805
22.8 0.000898667029377483
22.9 0.000809805951367347
23 0.000911801767471536
23.1 0.000772987156153566
23.2 0.000477697557820781
23.3 0.000621723774102673
23.4 0.000502408801737035
23.5 0.000357214728462955
23.6 0.000635341186399885
23.7 0.000708239961005285
23.8 0.000762649638474294
23.9 0.000799934839281067
24 0.00100875594907718
24.1 0.00092373611982632
24.2 0.000713370648012384
24.3 0.000670016054757896
24.4 0.000725215970362317
24.5 0.000773678889085173
24.6 0.000833851000474213
24.7 0.000703661225977375
24.8 0.000733109271269768
24.9 0.000662343736118036
25 0.000616055991436885
25.1 0.000528064309980171
25.2 0.000843626052030132
25.3 0.000582707419971831
25.4 0.000710313997601959
25.5 0.000870970078469133
25.6 0.000623217066965274
25.7 0.000818480109092624
25.8 0.000974934826186047
25.9 0.00072314818249674
26 0.00054684077052434
26.1 0.000813097302476729
26.2 0.000572341583729559
26.3 0.000456125882261828
26.4 0.000462274464398028
26.5 0.000420423730109992
26.6 0.000564418597401537
26.7 0.000418416490197651
26.8 0.000554565646367333
26.9 0.000535733010018129
27 0.000406891273923051
27.1 0.000427969118694948
27.2 0.000504302439952392
27.3 0.00151345106125084
27.4 0.000750185351145669
27.5 0.00124593247517885
27.6 0.000911737847670965
27.7 0.000956417165105982
27.8 0.000652922516071572
27.9 0.000836827092230456
};
\addplot [line width = \linewidthErrorStdVar, color = Koopman] 
table {%
0 0
0.1 0.00231382155809625
0.2 0.00174986460561837
0.3 0.000521473999119349
0.4 0.000991335148159158
0.5 0.0011123898652819
0.6 0.000826073506929425
0.7 0.000830010117497585
0.8 0.00118579388253674
0.9 0.00109054961755824
1 0.00092272141404117
1.1 0.000735840113387553
1.2 0.00116620234622212
1.3 0.00100038067579605
1.4 0.000930608722818219
1.5 0.000890634100684767
1.6 0.000874882364762989
1.7 0.00111799988969234
1.8 0.000500681534138414
1.9 0.000550055482228143
2 0.000629157126586424
2.1 0.000938304045859337
2.2 0.000533932167282896
2.3 0.000945623610745909
2.4 0.000919544215033855
2.5 0.000466442106561947
2.6 0.00122047429037796
2.7 0.0012105009133116
2.8 0.000486485945157758
2.9 0.000977131453425499
3 0.000613894969847019
3.1 0.000817722344022048
3.2 0.000903684503877579
3.3 0.000568164246540783
3.4 0.000856107706360912
3.5 0.000558153181636606
3.6 0.00126359047389572
3.7 0.00114533691041616
3.8 0.000874080601113791
3.9 0.00093454598721081
4 0.00103476348650648
4.1 0.000624812344385531
4.2 0.000723151201333599
4.3 0.00108206762044156
4.4 0.00115242509184273
4.5 0.000509198290058789
4.6 0.00110380351125568
4.7 0.000970567109691654
4.8 0.00133973966453398
4.9 0.000559141977077853
5 0.000597800514839613
5.1 0.00163858340003746
5.2 0.000792414061259948
5.3 0.000919313893704079
5.4 0.00109598962196249
5.5 0.000749247100236153
5.6 0.000630490658182267
5.7 0.000697438100003723
5.8 0.000835914006116052
5.9 0.000716131693396821
6 0.00128902088055644
6.1 0.000774245591486976
6.2 0.000988878122143022
6.3 0.00113345094009276
6.4 0.00128873180672814
6.5 0.00144432310637114
6.6 0.00166927755457942
6.7 0.00173019436296492
6.8 0.00128291320360566
6.9 0.00113998991396086
7 0.000918835658622885
7.1 0.000679578542263456
7.2 0.000494312204127911
7.3 0.000450104258328163
7.4 0.000334947693737067
7.5 0.00233697149816379
7.6 0.00185552855490247
7.7 0.000420624822008737
7.8 0.000927546068707483
7.9 0.000913894837169301
8 0.000830914025253391
8.1 0.000992114342055241
8.2 0.00119893235420779
8.3 0.000844849186910553
8.4 0.00097157893006511
8.5 0.000876176546543679
8.6 0.000979884411707082
8.7 0.000541959659784739
8.8 0.000552737010915882
8.9 0.000981541763191271
9 0.000972605996428251
9.1 0.000786760301073484
9.2 0.000908656885660565
9.3 0.000954639142338446
9.4 0.000839342969118711
9.5 0.000986830749668205
9.6 0.000901500198226887
9.7 0.00122096069891553
9.8 0.001035423542903
9.9 0.000738657930110164
10 0.00103897941989842
10.1 0.00123339343960432
10.2 0.00117879152336734
10.3 0.00126013673036793
10.4 0.000974661923062277
10.5 0.0010184098526295
10.6 0.00128482964123976
10.7 0.00105129832221835
10.8 0.000968090669500494
10.9 0.00109647353713529
11 0.00103326065460577
11.1 0.000930256850486395
11.2 0.000672720938816891
11.3 0.000992094092460991
11.4 0.000875099242841079
11.5 0.000987518477205598
11.6 0.000784275225117406
11.7 0.00135456697713248
11.8 0.00125993897128347
11.9 0.000753105525091796
12 0.000675029777917687
12.1 0.000705358927470156
12.2 0.00114772775785879
12.3 0.00114092390052541
12.4 0.00107677923464777
12.5 0.00198479136163936
12.6 0.00200495676479348
12.7 0.00267455378242174
12.8 0.00252522101137611
12.9 0.00250428334377271
13 0.00200298209420446
13.1 0.00140775671622728
13.2 0.00169670188171006
13.3 0.00173407531352689
13.4 0.00134607234503036
13.5 0.0010955544101341
13.6 0.00123446299106776
13.7 0.00135578971820703
13.8 0.00158616425942461
13.9 0.00181866400998842
14 0.00187715461462363
14.1 0.00193825994187203
14.2 0.00144310362383119
14.3 0.00127941593597263
14.4 0.00105268301433719
14.5 0.000814022211225082
14.6 0.000622488821800507
14.7 0.000641586439733255
14.8 0.000314938721211812
14.9 0.00247850102864788
15 0.00223214359343289
15.1 0.00102844243315233
15.2 0.00147051601771364
15.3 0.00197454569100926
15.4 0.00174461916787
15.5 0.00111925423496654
15.6 0.00149783053863594
15.7 0.00125952442926746
15.8 0.000862842844598773
15.9 0.00110250646726383
16 0.00121539660201696
16.1 0.000830968613504923
16.2 0.000722503710457852
16.3 0.000847761675673719
16.4 0.000570058875205066
16.5 0.00106452387915394
16.6 0.000601459590367437
16.7 0.00120254834380059
16.8 0.000794470229155346
16.9 0.000937184902292423
17 0.000882883052882641
17.1 0.000660592404660236
17.2 0.000570500660050605
17.3 0.000644114727357958
17.4 0.000550791963056205
17.5 0.000564626037781922
17.6 0.000962605350165296
17.7 0.000560863123892184
17.8 0.00100356340686651
17.9 0.000958999150272623
18 0.000919675335132375
18.1 0.00120269890371762
18.2 0.00128748430524133
18.3 0.000606464433023574
18.4 0.000533676637588932
18.5 0.000614330750395674
18.6 0.000875093933570785
18.7 0.000952594970414312
18.8 0.000859723338646776
18.9 0.000851478021904575
19 0.000607520763651768
19.1 0.000607153500922389
19.2 0.000585628426338073
19.3 0.000930621225656116
19.4 0.00054848005075443
19.5 0.000999132965025807
19.6 0.000987622520888055
19.7 0.00115768499120896
19.8 0.000760857848185578
19.9 0.00217868443749937
20 0.0025757532457817
20.1 0.00207432134123099
20.2 0.00142866174139104
20.3 0.00254784611964488
20.4 0.00184142839867851
20.5 0.000853156570506987
20.6 0.00109232285508926
20.7 0.0012973481281092
20.8 0.00155508978655704
20.9 0.000861571217779736
21 0.00093263536608656
21.1 0.00118309981572182
21.2 0.00125364798566768
21.3 0.00156947201832797
21.4 0.00170204760274928
21.5 0.00171366580158903
21.6 0.00130483811482201
21.7 0.00118086179402456
21.8 0.000931936017224753
21.9 0.000664528382793357
22 0.000490760807422744
22.1 0.000281347199176296
22.2 0.000191017192758325
22.3 0.00236696155268246
22.4 0.00236328733502591
22.5 0.00187616405394143
22.6 0.00192645536123725
22.7 0.00253863459847572
22.8 0.00179965746473059
22.9 0.00130425785357984
23 0.00151577686891308
23.1 0.00108714647039781
23.2 0.00070834692040687
23.3 0.00121358785845256
23.4 0.000888372037969257
23.5 0.000550482608614709
23.6 0.00112809963945043
23.7 0.00114982277362499
23.8 0.00129473823190493
23.9 0.00148186688506181
24 0.00190206407880203
24.1 0.00176082792094983
24.2 0.00110483423312575
24.3 0.000999349114697908
24.4 0.000998093868544542
24.5 0.00124792654650288
24.6 0.0013920839234567
24.7 0.00111135082035635
24.8 0.00115880031633385
24.9 0.000926167006732821
25 0.00105521472216144
25.1 0.000868361381014774
25.2 0.00153992767564172
25.3 0.000919609780331292
25.4 0.00101711837589698
25.5 0.00135878133609676
25.6 0.000923107221516305
25.7 0.00139683838152107
25.8 0.00161383105905065
25.9 0.00118038272201498
26 0.000818681400112886
26.1 0.00136885193230068
26.2 0.00101882430201544
26.3 0.000691494193709063
26.4 0.000686111480391842
26.5 0.000709339391666605
26.6 0.00101982176323454
26.7 0.00068081109359308
26.8 0.000974395440697961
26.9 0.00103464579775581
27 0.000619928796904854
27.1 0.000690288691454762
27.2 0.000812848642398674
27.3 0.001957096074384
27.4 0.00159860627015604
27.5 0.00243437390615917
27.6 0.00214864291646018
27.7 0.00193049644072864
27.8 0.000830450552688482
27.9 0.00191806377103793
};
\end{axis}
\end{tikzpicture}

%% file: figures/tikz/lessdata_b1_new.tex
\begin{tikzpicture}%
\begin{axis}[%
width = .38\textwidth,
height = \heightErrorN,
at={(0, 0)},
legend cell align={left},
legend columns = 3,
legend style={
  fill opacity=1,
  draw opacity=1,
  text opacity=1,
  at={(0.2,1.05)},
  anchor=south west,
  column sep = 0.25cm
},%
xlabel = {time (s)}, 
xmin=-0, xmax=3,
ymin=0.000145108448524775, ymax=0.233723638032451,
ymode=log,
]%
\addplot [line width = \lineWidthErrorN, color = RungeKuttaError]%
table {%
0 0
0.1 0.0165135818582296
0.2 0.0173573550131051
0.3 0.0155240468737607
0.4 0.016691397105289
0.5 0.0143597427283743
0.6 0.0143277101587927
0.7 0.0141366060316006
0.8 0.0132617453957916
0.9 0.0128695136926716
1 0.0124076789840141
1.1 0.0126642841967782
1.2 0.0119596603709604
1.3 0.0118711563344089
1.4 0.0111805312022035
1.5 0.0107251605082315
1.6 0.0105427094466726
1.7 0.0105103942337942
1.8 0.00985982763058332
1.9 0.0103415019491021
2 0.00961337937132115
2.1 0.009176374240606
2.2 0.00801289848433475
2.3 0.00777665982657768
2.4 0.0080071078808638
2.5 0.0089096573833126
2.6 0.0113768144136755
2.7 0.00947614752641055
2.8 0.0089458000759625
2.9 0.00891442844297832
3 0.0110135746945919
};
\addlegendentry{first principles}%
\addplot [line width = \lineWidthErrorN, color = n1]%
table {%
0 0
0.1 0.0168547775726986
0.2 0.00148356996263524
0.3 0.00169140700049326
0.4 0.00567886035490619
0.5 0.00506867470821116
0.6 0.00172323321283047
0.7 0.0015779221033442
0.8 0.00416152555713289
0.9 0.00389955994174455
1 0.000798252110969428
1.1 0.00256938661463039
1.2 0.000373631108651846
1.3 0.00154784972300235
1.4 0.000576092227241258
1.5 0.00088091019309275
1.6 0.000264028222304142
1.7 0.00136843922082343
1.8 0.000609183899763228
1.9 0.00229624541683824
2 0.000478179499443402
2.1 0.000321676864699139
2.2 0.00166068600174685
2.3 0.000419506047125796
2.4 0.00250094497523073
2.5 0.00239357229708892
2.6 0.00444473103999932
2.7 0.00190549954896417
2.8 0.000333395126520599
2.9 0.00120727362785229
3 0.00328798935183676
};
\addlegendentry{$n = 1$}
\addplot [line width = \lineWidthErrorN, color = n20]
table {%
0 0
0.1 0.0107520046142247
0.2 0.00703012533295705
0.3 0.00914622109847614
0.4 0.0106978338302807
0.5 0.0080884136978301
0.6 0.00780451593980268
0.7 0.00782457410451835
0.8 0.00722562085079703
0.9 0.00891159770822631
1 0.00706118253863211
1.1 0.00759381786545972
1.2 0.00662928014698138
1.3 0.00695113504294955
1.4 0.00630578552458954
1.5 0.00642374990273454
1.6 0.00577892116526725
1.7 0.00619664180256355
1.8 0.00577753317339912
1.9 0.00601060039346849
2 0.00540674142439366
2.1 0.00544416750963064
2.2 0.00573971102864734
2.3 0.0048921279110506
2.4 0.00495427396964455
2.5 0.00444867188324502
2.6 0.00479381797469929
2.7 0.00590327777469894
2.8 0.00454789990000331
2.9 0.00399834964862631
3 0.00372148446618804
};
\addlegendentry{$n = 20$}%
\addplot [line width = \lineWidthErrorN, color = n50]%
table {%
0 0
0.1 0.0203182924000263
0.2 0.0113276944775643
0.3 0.011849451266052
0.4 0.00769612610285905
0.5 0.0174108700990504
0.6 0.0115449522486172
0.7 0.0118855032277033
0.8 0.0174574096169832
0.9 0.0103252520981193
1 0.0134727784348086
1.1 0.0119737329604545
1.2 0.0146000918001689
1.3 0.0133250697103859
1.4 0.0152715300774054
1.5 0.0144506685252092
1.6 0.0149753041238364
1.7 0.0141513867404197
1.8 0.0159680451492941
1.9 0.0136039863195457
2 0.0161461804984792
2.1 0.0156359940126427
2.2 0.0174030960440711
2.3 0.0161872730270556
2.4 0.0136759546937059
2.5 0.0135912132715082
2.6 0.0119497986521269
2.7 0.0173607996267529
2.8 0.0155252933484477
2.9 0.0145914329956792
3 0.012211805863653
};
\addlegendentry{$n = 50$}%
\addplot [line width = \lineWidthErrorN, color = n100]%
table {%
0 0
0.1 0.0933999981288589
0.2 0.0926791017161514
0.3 0.0902161505732473
0.4 0.0902438387460173
0.5 0.0825383448184232
0.6 0.0839399792925418
0.7 0.0818470967294791
0.8 0.0776295250010234
0.9 0.0798208064035864
1 0.0760140054005865
1.1 0.0752048315891476
1.2 0.0721635656360341
1.3 0.0713763125336862
1.4 0.0691227358920897
1.5 0.0676327364980061
1.6 0.0659148212444659
1.7 0.0645714371656367
1.8 0.0619536830491499
1.9 0.0607896154753264
2 0.0573091656383364
2.1 0.0552321375810049
2.2 0.0516499860037895
2.3 0.0492233768548466
2.4 0.0467331101025748
2.5 0.0436695235227793
2.6 0.0404854600722299
2.7 0.0390111508653
2.8 0.0379974974058325
2.9 0.038801886467359
3 0.0412726391072552
};
\addlegendentry{$n = 100$}%
\end{axis}%
\begin{axis}[%
width = .38\textwidth,
height = \heightErrorN,
at={(.31\textwidth, 0)},
legend cell align={left},
legend columns = 2,
legend style={
  fill opacity=1,
  draw opacity=1,
  text opacity=1,
  at={(1,1)},
  anchor=south east,
  column sep = 0.25cm
},
xlabel = {time (s)}, 
xmin=-0, xmax=3,
ymin=0.000145108448524775, ymax=0.233723638032451,
yticklabels = {},
ymode=log,
]%
\addplot [line width = \lineWidthErrorN, color = n1]%
table {%
0 0
0.1 0.0170069580861405
0.2 0.00140108710254583
0.3 0.00140166355322037
0.4 0.00524505512800558
0.5 0.00548379538129153
0.6 0.00131104362399937
0.7 0.00114401085311314
0.8 0.00458180649427533
0.9 0.00347765639867595
1 0.000393577703062499
1.1 0.00216158993567225
1.2 0.00051132053399086
1.3 0.00114818171993019
1.4 0.000950680860953724
1.5 0.0005169422718385
1.6 0.000254150965122212
1.7 0.00102315000557671
1.8 0.000968693191825949
1.9 0.00195887117092178
2 0.000765285060260696
2.1 0.000403483349547753
2.2 0.00194988024257816
2.3 0.000692710960952163
2.4 0.00222684337625356
2.5 0.00213746120278164
2.6 0.00417121081270337
2.7 0.00217147326910141
2.8 0.000282310699715259
2.9 0.000903478356395921
3 0.0031183003184554
};
\addplot [line width = \lineWidthErrorN, color = n20]%
table {%
0 0
0.1 0.016324256353984
0.2 0.00074922458318092
0.3 0.00202276598665036
0.4 0.0049800220301007
0.5 0.00588510757008295
0.6 0.00100287896277084
0.7 0.000964578089686334
0.8 0.00503298303191267
0.9 0.00310395844945226
1 0.000603396347684195
1.1 0.00170677595678879
1.2 0.00114498278106672
1.3 0.000831252902677732
1.4 0.00161108352750722
1.5 0.000681212381770717
1.6 0.000822934354859154
1.7 0.000662718346706381
1.8 0.00168515529805249
1.9 0.00132523769164896
2 0.00152087521345923
2.1 0.00105973505875112
2.2 0.00273167306878487
2.3 0.00146453098478398
2.4 0.00145657790291938
2.5 0.00131695625243886
2.6 0.00334100766291158
2.7 0.00300686523940952
2.8 0.00106200234801933
2.9 0.000410397904973214
3 0.00217766362144465
};
\addplot [line width = \lineWidthErrorN, color = n50]%
table {%
0 0
0.1 0.0172869416888037
0.2 0.00170236706334129
0.3 0.00115936929110603
0.4 0.0053263463032189
0.5 0.00544258401478348
0.6 0.00143192537455004
0.7 0.00120832232729362
0.8 0.00457674112606484
0.9 0.00349655148232675
1 0.00038333026982136
1.1 0.00216397541600475
1.2 0.000504354055491986
1.3 0.00108091732532949
1.4 0.00103254962259857
1.5 0.000355549301347473
1.6 0.000456801227495615
1.7 0.000824868786196053
1.8 0.00115783179291279
1.9 0.00167984667289002
2 0.0011099642095187
2.1 0.000650509355301443
2.2 0.00246572174851809
2.3 0.00130732071227655
2.4 0.00149285133885671
2.5 0.00123402791476715
2.6 0.00313382906577376
2.7 0.00337774788247286
2.8 0.00159590064532075
2.9 0.000891028175052442
3 0.00136466777763771
};
\addplot [line width = \lineWidthErrorN, color = n100]%
table {%
0 0
0.1 0.0772943787246779
0.2 0.0665255209801766
0.3 0.0575907266458193
0.4 0.0534691317695794
0.5 0.0381210714228748
0.6 0.0340932957962354
0.7 0.0264599696272329
0.8 0.0143969567419319
0.9 0.0161006729696299
1 0.0135212180707904
1.1 0.0195525981449623
1.2 0.0269650897918591
1.3 0.0367237923427581
1.4 0.0475616380025682
1.5 0.058880175717317
1.6 0.0712161881753133
1.7 0.0843582460620498
1.8 0.0989536781702462
1.9 0.113715127842012
2 0.130171352583139
2.1 0.148101330900094
2.2 0.167564727912161
2.3 0.188205592105269
2.4 0.209211921561547
2.5 0.231721016990809
2.6 0.254517103896584
2.7 0.281428636644663
2.8 0.309246509298716
2.9 0.337780221263813
3 0.367296165046544
};
\addplot [line width = \lineWidthErrorN, color = RungeKuttaError]%
table {%
0 0
0.1 0.0165135818582296
0.2 0.0173573550131051
0.3 0.0155240468737607
0.4 0.016691397105289
0.5 0.0143597427283743
0.6 0.0143277101587927
0.7 0.0141366060316006
0.8 0.0132617453957916
0.9 0.0128695136926716
1 0.0124076789840141
1.1 0.0126642841967782
1.2 0.0119596603709604
1.3 0.0118711563344089
1.4 0.0111805312022035
1.5 0.0107251605082315
1.6 0.0105427094466726
1.7 0.0105103942337942
1.8 0.00985982763058332
1.9 0.0103415019491021
2 0.00961337937132115
2.1 0.009176374240606
2.2 0.00801289848433475
2.3 0.00777665982657768
2.4 0.0080071078808638
2.5 0.0089096573833126
2.6 0.0113768144136755
2.7 0.00947614752641055
2.8 0.0089458000759625
2.9 0.00891442844297832
3 0.0110135746945919
};%
\end{axis}%
\end{tikzpicture}%